\documentclass[nohyperref]{article}

\usepackage{amsmath,amsfonts,bm}

\def\eqref#1{equation~\ref{#1}}

\def\1{\bm{1}}

\DeclareMathAlphabet{\mathsfit}{\encodingdefault}{\sfdefault}{m}{sl}
\SetMathAlphabet{\mathsfit}{bold}{\encodingdefault}{\sfdefault}{bx}{n}

\usepackage{microtype}
\usepackage{graphicx}
\usepackage{subfigure}
\usepackage{booktabs} 

\usepackage{hyperref}
\usepackage{pifont}

\usepackage[dvipsnames]{xcolor}
\definecolor{Tianlong_color}{rgb}{0.858, 0.188, 0.478}

\usepackage[accepted]{icml2022}

\usepackage{amsmath}
\usepackage{amssymb}
\usepackage{mathtools}
\usepackage{amsthm}
\usepackage{makecell}
\usepackage{diagbox}
\usepackage{subfigure}
\usepackage{multirow}
\usepackage{multicol}
\usepackage{varwidth}
\usepackage{adjustbox}
\usepackage{threeparttable}
\usepackage{enumitem}
\usepackage{algorithmic}
\renewcommand{\algorithmicrequire}{\textbf{Input:}}
\renewcommand{\algorithmicensure}{\textbf{Output:}}

\usepackage[capitalize,noabbrev]{cleveref}

\theoremstyle{plain}

\theoremstyle{definition}

\theoremstyle{remark}

\usepackage[textsize=tiny]{todonotes}

\icmltitlerunning{Coarsening the Granularity: Towards Structurally Sparse Lottery Tickets}

\begin{document}

\twocolumn[
\icmltitle{Coarsening the Granularity: Towards Structurally Sparse Lottery Tickets}



\icmlsetsymbol{equal}{*}

\begin{icmlauthorlist}
\icmlauthor{Tianlong Chen}{ut}
\icmlauthor{Xuxi Chen}{ut}
\icmlauthor{Xiaolong Ma}{neu}
\icmlauthor{Yanzhi Wang}{neu}
\icmlauthor{Zhangyang Wang}{ut}
\end{icmlauthorlist}

\icmlaffiliation{ut}{University of Texas at Austin}
\icmlaffiliation{neu}{Northeastern University}

\icmlcorrespondingauthor{Zhangyang Wang}{atlaswang@utexas.edu}

\icmlkeywords{Machine Learning, ICML}

\vskip 0.3in
]



\printAffiliationsAndNotice{}  

\begin{abstract}
The lottery ticket hypothesis (LTH) has shown that dense models contain highly sparse subnetworks (i.e., \textit{winning tickets}) that can be trained in isolation to match full accuracy. Despite many exciting efforts being made, there is one  ``commonsense" rarely challenged: a winning ticket is found by iterative magnitude pruning (IMP) and hence the resultant pruned subnetworks have only unstructured sparsity. That gap limits the appeal of winning tickets in practice, since the highly irregular sparse patterns are challenging to accelerate on hardware. Meanwhile, directly substituting structured pruning for unstructured pruning in IMP damages performance more severely and is usually unable to locate winning tickets. In this paper, we demonstrate \textbf{the first positive result} that a structurally sparse winning ticket can be effectively found in general. The core idea is to append ``post-processing techniques" after each round of (unstructured) IMP, to enforce the formation of structural sparsity. Specifically, we first ``re-fill" pruned elements back in some channels deemed to be important, and then ``re-group" non-zero elements to create flexible group-wise structural patterns. Both our identified channel- and group-wise structural subnetworks win the lottery, with substantial inference speedups readily supported by existing hardware. Extensive experiments, conducted on diverse datasets across multiple network backbones, consistently validate our proposal, showing that \textbf{the hardware acceleration roadblock of LTH is now removed}. Specifically, the structural winning tickets obtain up to $\{64.93\%, 64.84\%, 60.23\%\}$ running time savings at $\{36\%\sim 80\%, 74\%, 58\%\}$ sparsity on \{CIFAR, Tiny-ImageNet, ImageNet\}, while maintaining comparable accuracy. {\small Code is at \url{https://github.com/VITA-Group/Structure-LTH}}.
\end{abstract}

\section{Introduction}
\vspace{-1mm}
Recently, the machine learning research community has devoted considerable efforts and financial outlay to scaling deep neural networks (DNNs) to enormous sizes ($175$ billion parameter-counts in GPT-3~\citep{brown2020language}). Although such overparameterization simplifies the training of DNNs and dramatically improves their generalization~\citep{bartlett2021deep,du2018gradient,kaplan2020scaling}, it may severely obstruct the practical usage on resource-limited platforms like mobile devices, due to its large memory footprint and inference time~\citep{hoefler2021sparsity}. Pruning is one of the effective remedies that can be dated back to~\citet{lecun1990optimal}: it can eliminate substantial redundant model parameters and boost the computational and storage efficiency of DNNs.

\begin{figure}[t]
\centering
\includegraphics[width=1\linewidth]{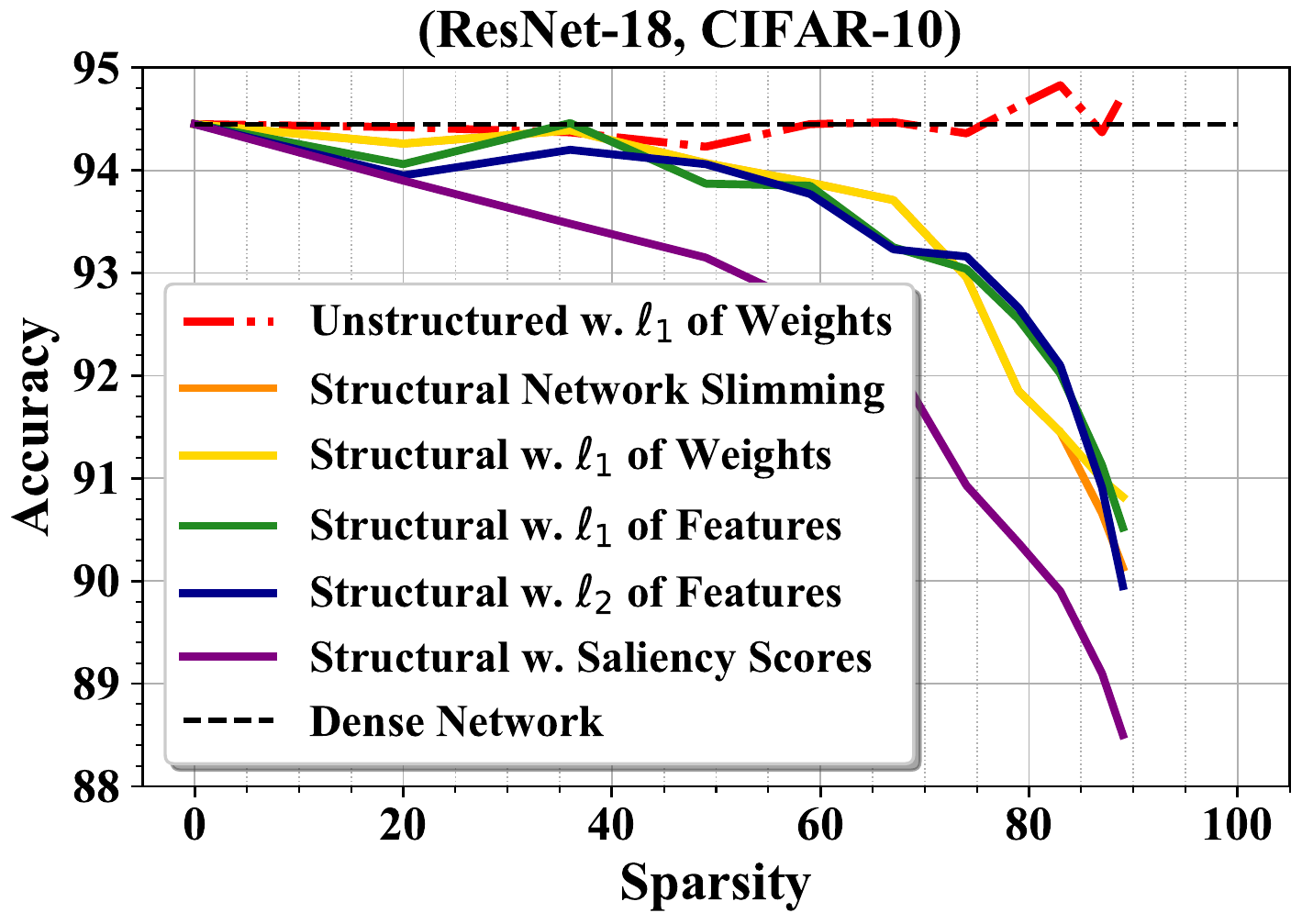}
\vspace{-8mm}
\caption{{\small Achieved test accuracy over different sparsity levels of diverse unstructured and structural subnetworks. Sparse models from classical channel-wise structural pruning algorithms~\citep{he2017channel,liu2017learning,bartoldson2019generalization,molchanov2019importance} can not match the full accuracy of the dense model.}}
\label{fig:tisser}
\vspace{-5mm}
\end{figure}

Such benefits drive numerous interests in designing model pruning algorithms~\citep{han2015deep,han2015learning,ren2018admmnn,he2017channel,liu2017learning}. Among this huge family, an emerging representative studies the prospect of training \textit{sparse subnetworks} in lieu of the full dense models without impacting performance~\citep{frankle2018the,chen2020lottery}. For instance, \citet{frankle2018the} demonstrates that dense models contain sparse, matching subnetworks~\citep{frankle2020linear} (a.k.a. \textit{winning tickets}) capable of training in isolation from the original initialization to match or even surpass the full accuracy. This phenomenon is referred to as the \textit{lottery tickets hypothesis} (LTH), which indicates several impressive observations: $(i)$ usually extreme sparsity levels (e.g., $90\%$, $95\%$) can be achieved without sacrificing the test accuracy; $(ii)$ the located winning ticket maintains undamaged expressive power as its dense counterpart, and can be easily trained from scratch or early-epoch weights~\citep{Renda2020Comparing,frankle2020linear} to recover the full performance. These advances are positive signs of the substantial potential of sparse DNNs. 

However, almost all LTH literature investigates unstructured sparsity only. In practical scenarios, it brings little hardware efficiency benefits due to the poor data locality and low parallelism~\citep{he2017channel,mao2017exploring,wen2016learning} caused by highly irregular sparse patterns. Meanwhile, most of the accelerators are optimized for dense matrix operations~\citep{han2016eie}, which means there is limited speedup for unstructured pruned subnetworks even if the sparsity level exceeds $95\%$~\citep{wen2016learning}. Structural pruning~\citep{he2017channel,liu2017learning} as an alternative to exploring sparse subnetworks, removes the entire filter or channel in DNNs to gain more computational efficiency at the cost of (more) accuracy degradation. As shown in Fig.~\ref{fig:tisser}, traditional channel-wise structural pruning approaches~\citep{he2017channel,bartoldson2019generalization,molchanov2019importance} quickly degrade performance and cannot lead to winning tickets, which was also echoed in~\citet{You2020Drawing}.

\vspace{-0.5mm}
In our paper, we present the first study into the \textit{structural lottery tickets}, which explores hardware-friendly structural sparsity (including channel-wise and group-wise patterns) in order to find lottery tickets. Specifically, we start from unstructured sparse subnetworks, and then adopt proposed \textit{refilling} techniques to create channel-wise structural sparsity by growing back the pruned elements within the most important channels and abandoning the rest. Our results (Section~\ref{sec:main_res}) show such refined channel-wise structural subnetworks win the lottery at a moderate sparsity level with $\sim 50\%$ running time savings on an Nvidia 2080 TI GPU. In order to push the compression ratio higher, we introduce a \textit{regrouping} algorithm based on hypergraph partitioning~\citep{rumi2020accelerating} to establish group-wise structural patterns which are more amenable to pruning due to the shape flexibility of grouped dense blocks. These group-wise structural winning tickets achieve $\sim60\%$ running time savings at $50\%\sim80\%$ sparsity without any performance degradation compared to the dense models. 

\vspace{-0.5mm}
Note that this paper focuses on \textit{general} structural sparse patterns capable of acceleration, including conventional channel-wise sparsity and other fine-grained structural sparsity. The latter actually becomes prevailing recently since it achieves superior performance and maintains satisfied speedup, sparking great interest in industries such as NVIDIA (N:M)~\citep{zhou2021learning} and Google (Block-wise)~\citep{shangguan2019optimizing}. Meanwhile, unlike~\citet{zhou2021learning}, our group-wise sparse patterns do 
NOT need any specific hardware accelerators and are generally applicable to common GPU devices. Lastly, although we mainly investigate inference efficiency, our proposals can also enable efficient training in transfer learning paradigms as demonstrated in Appendix~\ref{sec:more_results}. Our main contributions lie in the following aspects:
\vspace{-1mm}
\begin{itemize}
    \item To our best knowledge, we are the first to demonstrate the existence of structurally sparse winning tickets at non-trivial sparsity levels (i.e., $>30\%$), and with both channel-wise and group-wise sparse patterns. 
    \item We propose the \textit{refilling} technique and introduce the \textit{regrouping} algorithm to form channel-wise and group-wise structural sparsity. Such refined structural subnetworks match the trainability and expressiveness of dense networks, while enabling the inference speedup on practical hardware platforms like GPU machines (general and not tied to particular hardware).
    \item Extensive experiments validate our proposal on diverse datasets (i.e., CIFAR-10/100, Tiny-ImageNet, and ImageNet) across multiple network architectures, including ResNets, VGG, and MobileNet. Specifically, our structural winning tickets achieve $53.75\%\sim64.93\%$ GPU running time savings at $45\%\sim80\%$ channel- and group-wise sparsity.
\end{itemize}
\vspace{-1mm}

\vspace{-1mm}
\section{Related Work}
\vspace{-1mm}

\begin{figure*}[t]
\centering
\vspace{-0.5em}
\includegraphics[width=0.95\linewidth]{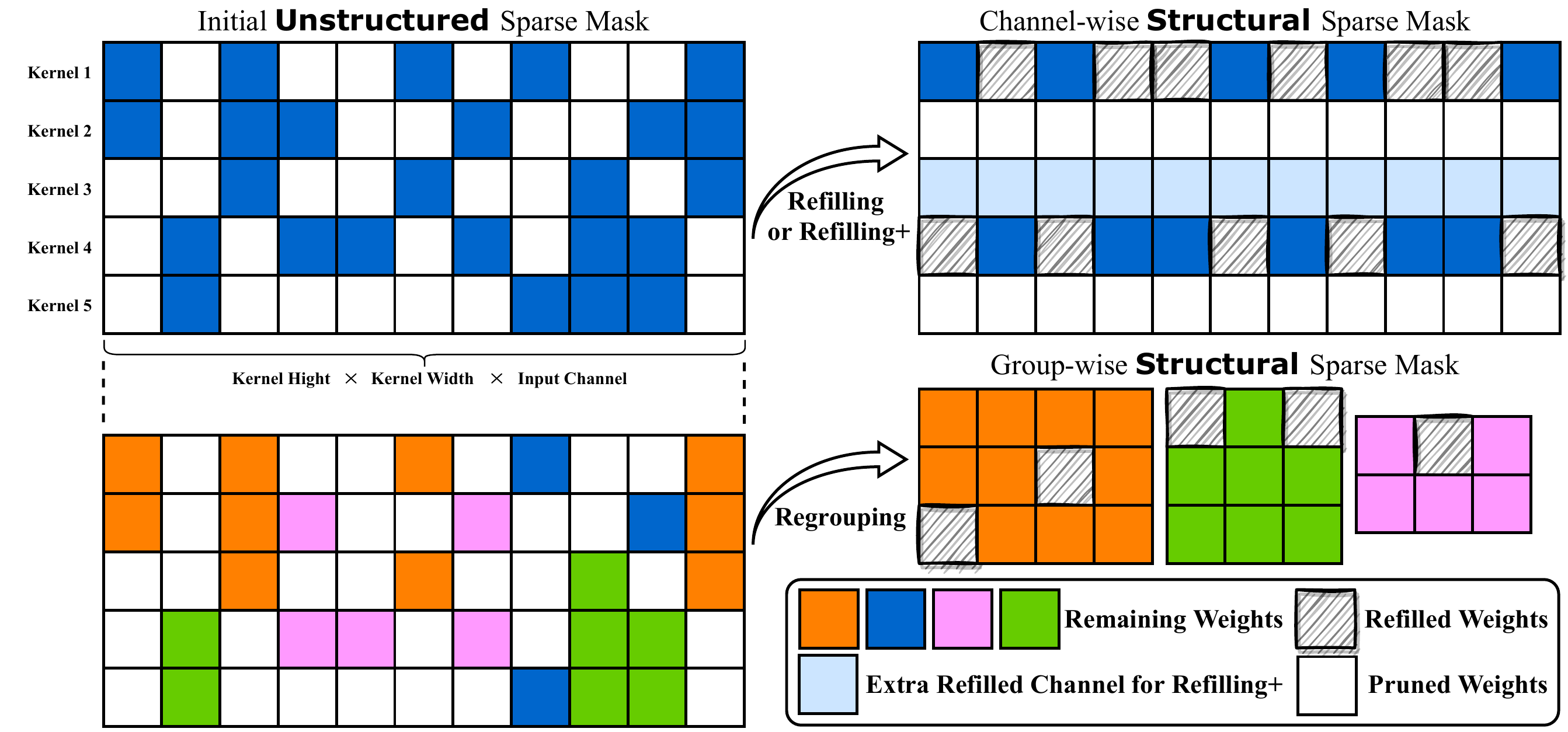}
\vspace{-4mm}
\caption{{\small Overview of our proposals including refilling, refilling+, and regrouping, which turn unstructured sparse mask into channel-wise and group-wise structured sparse masks.}}
\vspace{-5mm}
\label{fig:methods}
\end{figure*}

\paragraph{Pruning.} Network pruning is a technique that aims at eliminating the unnecessary model parameters~\citep{blalock2020state}, which can effectively shrink models for the deployment on resource-constrained devices~\citep{lecun1990optimal,hanson1988comparing}. Pruning algorithms are roughly categorized into two groups: (1) unstructured pruning~\citep{lecun1990optimal,han2015deep,han2015learning,ren2018admmnn,Zhang_2018} with irregular sparse patterns; (2) structural pruning~\citep{he2017channel,liu2017learning,li2016pruning,hu2016network,wen2016learning,hong2018efficient} with structural sparse patterns such as layer-wise, channel-wise, block-wise, column-wise, etc.. 

Within the group of unstructured pruning methods, \citet{han2015deep,han2015learning} remove insignificant connections of models in the post-training stage, with respect to certain heuristics like weight/gradient magnitudes; during training sparsification is also another popular trend for pruning by leveraging $\ell_0$ regularization~\citep{louizos2017learning} or alternating direction method of multipliers (ADMM)~\citep{ren2018admmnn,Zhang_2018}. Recently, several pruning-at-initialization methods~\citep{Wang2020Picking,snip,tanaka2020pruning} are proposed to identify critical unstructured connections for gradient-flow preserving, without any training. Although the unstructured sparse model has superior performance, it usually suffers from poor data locality and low parallelism~\citep{he2017channel,mao2017exploring,wen2016learning}, which make it hard to be accelerated in real-world hardware platforms. 

On the contrary, structural pruning is more hardware-friendly at the cost of notable accuracy loss when the compression ratio increases. \citet{he2017channel,liu2017learning} slim the network channels via $\ell_1$ regularization, and~\citet{bartoldson2019generalization} selects important channels according to heuristics of feature maps. To combine the benefits of structural and unstructured pruning, hybrid pruning strategies have been introduced to pursue more general structural spares patterns which are also capable of acceleration. For example, convolution kernels with half regular sparsity~\citep{chen2018sc} or pattern-based structural sparsity~\citep{ma2020pconv} or vector-wise~\citep{Zhu2019STC} and group-wise~\citep{rumi2020accelerating} regular sparsity.

\vspace{-1mm}
\paragraph{The lottery tickets hypothesis (LTH).} The lottery ticket hypothesis (LTH)~\citep{frankle2018the} conjectures that there exists a sparse subnetwork called winning ticket within a dense network, whose performance can match with the dense network when training from the same initialization. With the assistance of weight rewinding techniques~\citep{Renda2020Comparing,frankle2020linear}, the original LTH can be scaled up to larger networks and datasets. The existence of winning tickets are broadly verified under diverse contexts, such as image classification~\citep{frankle2018the,pmlr-v139-zhang21c,chen2020lottery2,ma2021good,gan2021playing,chen2021you}, natural language processing~\cite{gale2019state,chen2020lottery}, generative adversarial networks~\cite{chen2021gans,chen2021ultra}, graph neural networks~\cite{chen2021unified}, and reinforcement learning~\cite{yu2019playing}. However, all of the above LTH literature only locate \textit{unstructured} sparse winning tickets, which can hardly bring hardware efficiency boost to real-world applications.

As the most related work, \citet{You2020Drawing} finds structural winning tickets at only low sparsity levels around $30\%$ in a few cases. It again reveals the complication and difficulty of identifying computation-friendly sparse patterns. Another concurrent work~\citep{alabdulmohsin2021generalized} investigates a generalized LTH with weight space factorization, which is orthogonal to our work. 

\begin{table*}[t]
\centering
\vspace{-4mm}
\caption{Implementation details which follow the standard settings in~\citet{ma2021sanity}.}
\label{tab:all_exp}
\begin{adjustbox}{width=1\textwidth}
\begin{threeparttable}
\begin{tabular}{l|cccccccccc}
\toprule
\multirow{2}{*}{Settings} & \multicolumn{4}{c}{CIFAR-10} & \multicolumn{4}{c}{CIFAR-100} & \multicolumn{1}{c}{Tiny-ImageNet} & \multicolumn{1}{c}{ImageNet}\\ \cmidrule(lr){2-5} \cmidrule(lr){6-9} \cmidrule(lr){10-10} \cmidrule(lr){11-11} 
& WRN-32-2 & RN-18 & MBNet-v1 & VGG-16 & WRN-32-2 & RN-18 & MBNet-v1 & VGG-16 & RN-50 & RN-50\\ \midrule
Batch Size & 128 & 128 & 128 & 128 & - & - & 64 & - & 32 & - \\ \midrule
Weight Decay & \multicolumn{1}{c}{$1\times10^{-4}$} & $1\times10^{-4}$ & $1\times10^{-4}$ & \multicolumn{1}{c}{$2\times10^{-4}$} & $2\times10^{-4}$ & $2\times10^{-4}$ & $2\times10^{-4}$ & \multicolumn{1}{c}{$5\times10^{-4}$} & $5\times10^{-4}$ & $1\times10^{-4}$\\ \midrule
\multirow{1}{*}{Learning Rate} & \multicolumn{9}{c}{0.1;$\times0.1$ at 80,120 epoch of total 160 epochs} & 0.1;$\times0.1$ at 30,60 epoch of total 95 epochs\\  \midrule
Optimizer & \multicolumn{10}{c}{SGD~\citep{ruder2016overview} with a momentum of 0.9}\\ \midrule
Model Size & $1.86$ M & $11.22$ M & $3.21$ M & $14.72$ M & $1.86$ M & $11.22$ M & $3.21$ M & $14.72$ M & $25.56$ M & $25.56$ M \\
\bottomrule
\end{tabular}
\end{threeparttable}
\end{adjustbox}
\vspace{-4mm}
\end{table*}

\vspace{-3mm}
\paragraph{Sparse convolutional neural network acceleration on GPU.} Previous works have explored the acceleration of sparse convolution operations in two different directions. \underline{One direction} is to design efficient implementation of unstructured pruned networks for improved data locality and utilization of hardware~\citep{chen2018escort,park2016faster}. For example, \citet{dong2019acorns} proposes ``Acorns" to accelerate the sparse computations of convolution kernels with an input sparsity. \citet{peng2017adaptive} has proposed a matrix splitting algorithm for efficient inference of convolutional neural networks (CNN). Nvidia's cuSPARSE\footnote{\scalebox{0.66}{\url{https://docs.nvidia.com/cuda/archive/10.2/cusparse/index.html}}} library contains various efficient sparse matrix computation algorithms like SpMM on GPUs, drawing great attention to efficient scientific computing. Furthermore, advanced approaches are developed based on SpMM, such as Adaptive Sparse Tiling (ASpT)~\citep{hong2019adaptive}. ASpT significantly improves the data usage of SpMM and achieves the current state-of-the-art performance among SpMM implementation variants. \underline{Another direction} focuses on more hardware-friendly pruning methods~\citep{chen2018sc,ma2020pconv,niu2020patdnn}. During the model pruning, these works aim to maintain certain regular sparse patterns, which benefit the hardware processing/computing of corresponding sparse matrices. However, \citet{chen2018sc} achieves unsatisfactory compression ratio, while the pruning methods used in~\citet{ma2020pconv} and \citet{niu2020patdnn} require dedicated compiler optimization to accelerate network execution.

\vspace{-1mm}
\section{Methodology}
\subsection{Notations and Preliminaries} \label{sec: preliminaries}
\paragraph{Sparse subnetworks and pruning methods.} In this paper, we mainly follow the routine notations in~\cite{frankle2018the, Renda2020Comparing}. For a network $f(x;\theta)$ with input samples $x$ and model parameters $\theta$, a sparse subnetwork is a network $f(x;m\odot\theta)$ with a binary pruning mask $m\in\{0,1\}^{|\theta|}$, where $\odot$ is the element-wise product. In other words, it is a copy of dense network $f(x;\theta)$ with some weights fixed to $0$. If the non-fixed remaining weights are distributed irregularly, we call it \textbf{unstructured} sparse patterns (e.g., the \textit{left} of Figure~\ref{fig:methods}); if they are clustered into channels or groups, we name it \textbf{structural} sparse patterns (e.g., the \textit{right} of Figure~\ref{fig:methods}). 

To obtain the desired sparse subnetworks, we consider and benchmark multiple classical pruning algorithms: (1) \textit{random pruning} (\texttt{RP}) which usually works as a necessary baseline for the sanctity check~\citep{frankle2018the}; (2) \textit{one-shot magnitude pruning} (\texttt{OMP}) by eliminating a part of model parameters with the globally smallest magnitudes~\citep{han2015deep}; (3) \textit{the lottery ticket hypothesis}~\citep{frankle2018the} with iterative weight magnitude pruning (\texttt{LTH-IMP} or \texttt{IMP} for simplicity)~\citep{han2015deep}. As adopted in LTH literature~\citep{frankle2018the}, we identify the sparse lottery tickets by iteratively removing the $20\%$ of remaining weight with the globally smallest magnitudes, and rewinding model weights to the original random initialization~\citep{frankle2018the} or early training epochs~\citep{Frankle2020The,chen2020lottery2}. In this paper, the model weights are rewound to the eighth epoch (i.e., the $5\%$ of the entire training process) for all CIFAR, Tiny-ImageNet, and ImageNet experiments. (4) \textit{pruning at initialization} mechanisms. We choose several representative approaches such as \texttt{SNIP}~\citep{lee2018snip}, \texttt{GraSP}~\citep{Wang2020Picking}, and \texttt{SynFlow}~\citep{tanaka2020pruning}, which explore sparse patterns at random initialization with some gradient flow-based criterion. (5) \textit{Alternating Direction Method of Multipliers} (\texttt{ADMM}) for punning. It is a well-known optimization-based pruning method~\citep{niu2020patdnn,Zhang_2018}, which can obtain superior compression ratios with little performance degradation for deep neural networks. Note that all pruning approaches are mainly conducted over networks without counting their classification heads~\citep{frankle2018the}.

\vspace{-1mm}
\paragraph{Structural winning tickets.} We begin by extending the original lottery tickets hypothesis to the context of structural sparse patterns. A subnetwork $f(x;m\odot\theta)$ is a structural winning ticket for an algorithm $\mathcal{A}^{\mathcal{T}}_t$ if it satisfies: \ding{172} training subnetworks $f(x;m\odot\theta)$ with algorithm $\mathcal{A}_t^{\mathcal{T}}$ results in performance measurement on task $\mathcal{T}$ no lower than training dense networks $f(x;\theta)$ with algorithm $\mathcal{A}^{\mathcal{T}}_t$, where $\theta$ is the original random initialization $\theta_0$ or early rewound weights like $\theta_{5\%}$, and $t$ is the training iterations; \ding{173} the non-zero elements in pruning mask $m$ are clustered as channels, groups or other hardware-friendly structural patterns. 

\vspace{-1mm}
\paragraph{Implementation details.} We conduct experiments on diverse combinations of network architectures and datasets. Specifically, we adopt Wide-ResNet-32-2~\citep{zagoruyko2016wide} (or WRN-32-2), ResNet-18~\citep{he2016deep} (or RN-18), MobileNet-v1 (or MBNet-v1)~\citep{howard2017mobilenets}, and VGG-16~\citep{simonyan2014very} on both CIFAR-10~\citep{krizhevsky2009learning} and CIFAR-100 datasets. ResNet-50 (or RN-50) is evaluated on both Tiny-ImageNet~\citep{le2015tiny} and ImageNet~\citep{deng2009imagenet} datasets. Table~\ref{tab:all_exp} includes more training and evaluation details of our experiments. 

\subsection{Refilling for Structural Patterns}
\vspace{-1mm}
It is well-known that the irregular sparsity patterns from unstructured magnitude pruning block the acceleration on practical hardware devices. To overcome the limitation, we propose a simple \textit{refilling} strategy to reorganize the unstructured sparse patterns and make them more hardware friendly. Specifically, we \underline{first} select important channels from the unstructured subnetwork according to certain criteria. The number of picked channels are depended on the desired sparsity level. \underline{Then}, the pruned elements are grown back to be trainable (i.e., unpruned) and are reset to the same random initialization or early rewound weights. \underline{Lastly}, the rest parameters in the remaining insignificant channels will be removed. In this way, we refill important channels and empty the rest to create a channel-wise structural sparse pattern that essentially brings computational reductions. Note that the picking criterion can be the number of remaining weights in the channel, or the channel's weight statistics or feature statistics or salience scores, which are comprehensively investigated in the ablation (Section~\ref{sec:more_results}). The complete pipeline and illustration are summarized in Algorithm~\ref{alg:IMP_Refill} and Figure~\ref{fig:methods}, respectively.

\vspace{-4mm}
\begin{minipage}{0.48\textwidth}
\begin{algorithm}[H]
\caption{\texttt{IMP} with rewinding step $i$} \label{alg:IMP}
\begin{algorithmic}[1]
\REQUIRE {$f(x;\theta_0)$, unstructured sparsity $s$}
\ENSURE {$f(x; m\odot\theta_i)$}
\STATE Set the pruning mask $m=\boldsymbol{1}\in\mathbb R^{|\theta|}$\\
\STATE Train $f(x;\theta_0)$ for $i$ steps: $f(x;\theta_i)=\mathcal{A}_i^{\mathcal{T}}(f(x;\theta_0))$  \\
\WHILE {not reach sparsity $s$}
\STATE Train $f(x;m\odot\theta_i)$ for $t-i$ steps: $f(x;m\odot\theta_t)=\mathcal{A}_{t-i}^{\mathcal{T}}(f(x;m\odot\theta_i))$ \\
\STATE Pruning $20\%$ of remaining weight of $m\odot\theta_t$, and update $m$ 
\ENDWHILE 
\end{algorithmic}
\end{algorithm}
\end{minipage}
\vspace{-6mm}

\begin{minipage}{0.48\textwidth}
\begin{algorithm}[H] 
\caption{\texttt{IMP-Refill(+)}} \label{alg:IMP_Refill}
\begin{algorithmic}[1]
\REQUIRE {$f(x;m\odot\theta_i)$ with unstructured sparsity $s$ (Algo.~\ref{alg:IMP})}
\ENSURE {$f(x;m\odot\theta_i)$ with channel-wise structural mask $m$ at sparsity $\tilde{s}$}
\STATE Calculate importance scores of each channel according to certain criterion \\
\STATE Pick top-$k$ channels in $m$, refill back their $0$ (pruned) elements with $1$ (trainable) and update $m$, maintaining $\tilde{s}\sim s$ \\
\STATE Pick and refill back extra channels in $m$ with $\tilde{s}^+< s$ \\ 
\textcolor{gray}{\# Optional for \texttt{Refill+}}
\end{algorithmic}
\end{algorithm}
\end{minipage}
\vspace{-6mm}

\begin{minipage}{0.48\textwidth}
\begin{algorithm}[H] 
\caption{\texttt{IMP-Regroup}} \label{alg:regroup}
\begin{algorithmic}[1]
\REQUIRE {$f(x;m\odot\theta_i)$ with unstructured sparsity $s$ from Algorithm~\ref{alg:IMP}, hyperparameters $t_1$, $t_2$, $b_1$, and $b_2$}
\ENSURE {$f(x;m\odot\theta_i)$ with group-wise structural mask $m$ at sparsity $s^*$}
\WHILE {dense block can be found}
\STATE  {Divide the rows of the sparse pruning mask $m$ into $t_1$ groups using hypergraph partitioning (hMETIS)\footnote{\scalebox{0.75}{\url{http://glaros.dtc.umn.edu/gkhome/metis/hmetis/overview}}} \\
\FOR{group $c_i\in\{c_1,c_2,\dots,c_{t_1}\}$ }{
\IF{$c_i$ has $\ge b_1$ rows}{
\STATE Select columns in $c_i$ that has no less than $t_2$ non-zero items \\
\IF{$\ge b_2$ columns are selected}{
\STATE Group and Refill the selected columns as well as rows to a dense block, and update $m$ \\
}
\ENDIF
}
\ENDIF
}
\ENDFOR
}
\ENDWHILE
\STATE Set other elements out of dense blocks to $0$ \\
\end{algorithmic}
\end{algorithm}
\end{minipage}
\vspace{-4mm}

Here we provide a detailed description of how many and which channels we choose to refill. Our main experiments adopt the $\ell_1$ norm of channel weights as the picking criterion to score the channel importance due to its superior performance. Let $\theta^l\in\mathbb{R}^{c_{\mathrm{out}}\times n}$ denotes the parameters of the convolutional layer $l$, where $c_{\mathrm{out}}$ is the number of output channel and $n$ is the continued product of the number of input channel, channel height and weight, as shown in Figure~\ref{fig:methods}. $\theta^l_i\in\mathbb{R}^{n}$ represents the weights in the $i$th kernel and $m^l_i\in\{0,1\}^{|\theta^l_i|}$ is the corresponding mask. We first calculate the $\ell_1$ norm of $m^l_i\odot\theta^l_i$, which is a summation of the absolute value of remaining weights in the kernel $i$. Then we use it to pick the top-$k$ scored kernels, which will be fully refilled. $k=\lceil s^l\times c_{\mathrm{out}} \times n\rceil$, where $s^l$ is the original layerwise sparsity and $c_{\mathrm{out}} \times n$ is the total number of weights in kernel $i$. Meanwhile, the rest $c_\mathrm{out}-k$ kernels are dropped for efficiency gains.

Furthermore, we propose a soft version, \textit{refilling+}, to make a redemption for the aggressive nature of wiping out all remaining channels. It picks and re-actives an extra proportion of channels to slow down the network capacity reduction, as indicated by shallow blue blocks in Figure~\ref{fig:methods}.

\begin{figure*}[t]
\centering
\vspace{-0.5em}
\includegraphics[width=1\linewidth]{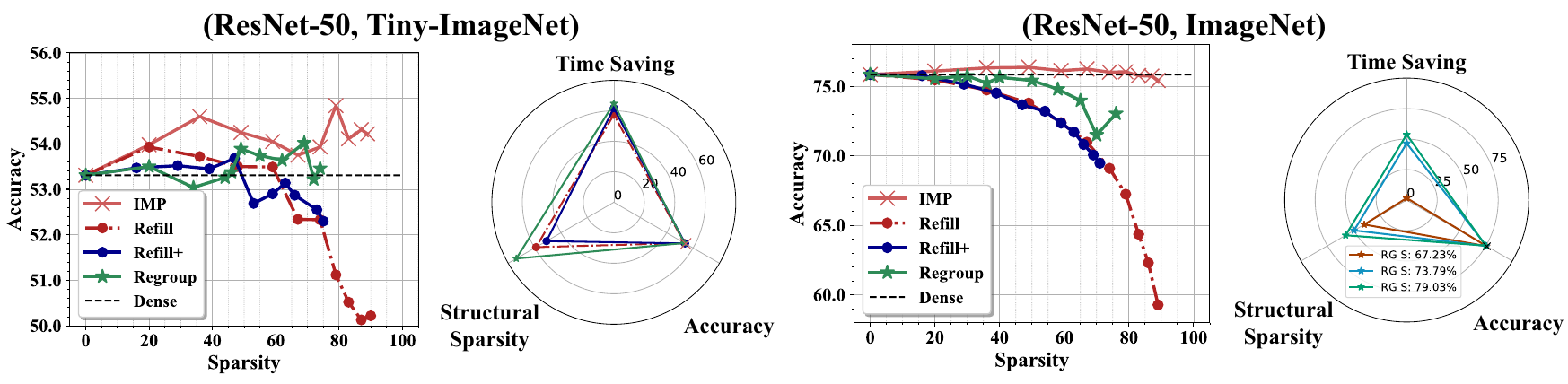}
\vspace{-9mm}
\caption{{\small (\textit{Curve plots}) Testing accuracy (\%) over network sparsity (\%) on Tiny-ImageNet and ImageNet datasets with ResNet-50 ($25.56$ M). (\textit{Radar plots}) The end-to-end inference time saving of extreme structural winning tickets. Unstructured subnetworks or dense models do not have structural sparsity, and thus they are plotted as dots in the axes of accuracy in the corresponding radar plot. The rightmost plot includes three extreme regroup tickets with accuracy drop $<1\%$, where ``RG S: $x\%$" indicates unstructured sparsity before regrouping.}}
\vspace{-0.7em}
\label{fig:imagenet_res}
\end{figure*}

\subsection{Regrouping for Structural Patterns}

Although proposed \textit{refilling+} reorganizes the unstructured mask and produces useful channel-wise structural subnetworks, it is rigid and inelastic since the smallest manageable unit is a kernel. In other words, the dense matrices in identified structural patterns have a restricted shape where one dimension must align with the kernel size $n$, i.e., the continued product of the number of input channels, channel height, and weight. Motivated by~\citet{rumi2020accelerating}, we introduce a \textit{regrouping} strategy (Figure~\ref{fig:methods}) to create more fine-grained group-wise structural patterns with flexible shapes for remaining dense matrices. 

$\rhd$ \textbf{How to perform regrouping?} \textit{Regrouping} aims to find and extract dense blocks of non-pruned elements in the sparse weight matrix. These blocks have diverse shapes, as demonstrated in Figure~\ref{fig:methods}, which are usually smaller in size compared to the original sparse matrix. Note that a channel/kernel can be regarded as a special case of the dense block. 

As described in Algorithm~\ref{alg:regroup}, to achieve the goal, we first need to find similar rows and columns, and then bring them together. Specifically, We adopt the Jaccard similarity~\citep{rumi2020accelerating,10.1145/3332466.3374546} among non-zero columns as the similarity between two rows in the sparse matrix, which is calculated as a cardinality ratio of the intersections to the union of non-zero columns. For instance, kernel $1$ and kernel $2$ in Figure~\ref{fig:methods} (upper left) share three columns in eight non-zero distinct columns, and their similarity is $\frac{3}{8}$. Then, if two rows have a larger similarity, they can form a denser block when we group them together. Take Figure~\ref{fig:methods} as an example. We can group kernel $1,2,3$'s non-zero columns $1,3,6,11$ with at least two elements together, which leads to the first orange dense block.

More precisely, we take the hypergraph partitioning in the regrouping algorithm to generate dense blocks. It treats each row and column from the sparse matrix as a node and hyperedge in the hypergraph, where hyperedge (i.e., column) connects the corresponding nodes (i.e., row). Then, the pair-wise similarity is leveraged to locate an optimal partitioning, which can be achieved with hMETIS\footnote{\scalebox{0.75}{\url{http://glaros.dtc.umn.edu/gkhome/metis/hmetis/overview}}}. More details are referred to~\citet{rumi2020accelerating}. After obtaining the desired dense blocks, we enable all their parameters to be trainable by refilling the corresponding pruned elements. Note that refilling these pruned weights does not cause any efficiency loss since the size of the blocks is fixed, while it potentially maximizes the usage of these blocks and brings accuracy gains. Meanwhile, the rest parameters not included in the dense blocks will be discarded, i.e., setting the corresponding position in binary mask $m$ to zero, for reducing the computational overhead as illustrated in Figure~\ref{fig:methods}. It is because any parameters outside the dense blocks require extra weights loading and have little data reuse~\citep{rumi2020accelerating}, which harms the trade-off of accuracy and efficiency.

$\rhd$ \textbf{How refilled / regrouped dense blocks be beneficial?} We notice that the common tools like cuDNN~\citep{chetlur2014cudnn} have a significant drawback that the inference time does not linearly change with the number of kernels, since they are only optimized for kernel matrices with a multiple of $32$ rows~\citep{radu2019performance}. For example, as stated in~\citet{rumi2020accelerating}, a convolutional layer with $10$ kernels might have a similar inference time with a convolutional layer with $32$ kernels. However, the number of kernels in these dense blocks is almost arbitrary, so a more sophisticated GEMM-based efficient implementation~\citep{rumi2020accelerating} is needed to accelerate better our refilled / regrouped structural patterns. Following~\citet{rumi2020accelerating}, we split a kernel with $r$ rows into two parts: one has $[r/32]\times 32$ rows and the other one has $r$ mod $32$ rows. First, we directly apply the standard GEMM-based convolution algorithm with shared memory to cache the input and output matrix. For the second part, due to the poor data reuse of input matrices, we choose to cache the kernel and output matrices for an improved cache hit rate and overall performance. More details are referred to~\citet{rumi2020accelerating}.

\begin{figure*}[t] 
\centering
\vspace{-0.5em}
\includegraphics[width=1\linewidth]{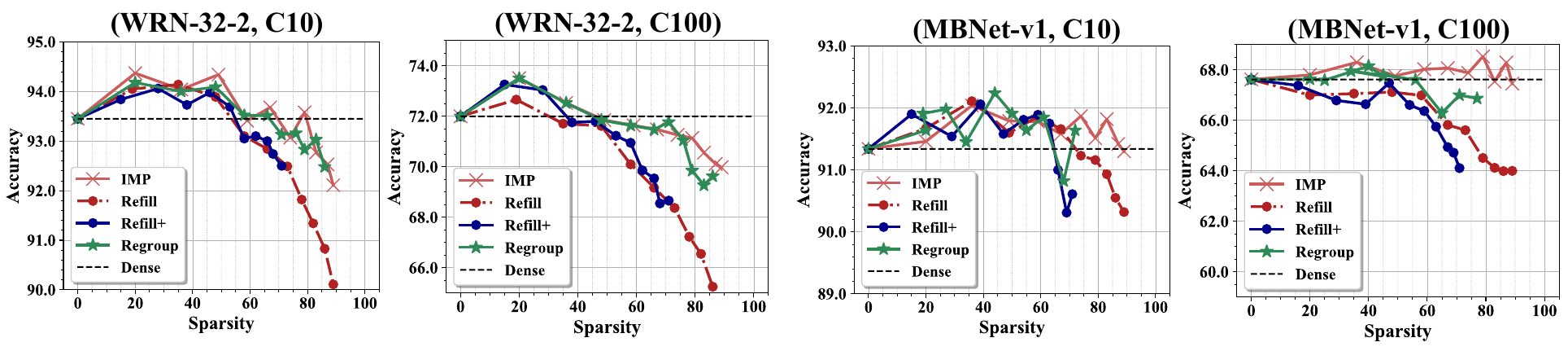}
\vspace{-8mm}
\caption{{\small Testing accuracy (\%) over sparsity (\%) on CIFAR-10/100 with Wide-ResNet-32-2 ($1.86$ M) and MobileNet-v1 ($3.21$ M).}}
\vspace{-0.4em}
\label{fig:cifar_res_small}
\end{figure*}

\begin{figure*}[t] 
\centering
\vspace{-0.5em}
\includegraphics[width=1\linewidth]{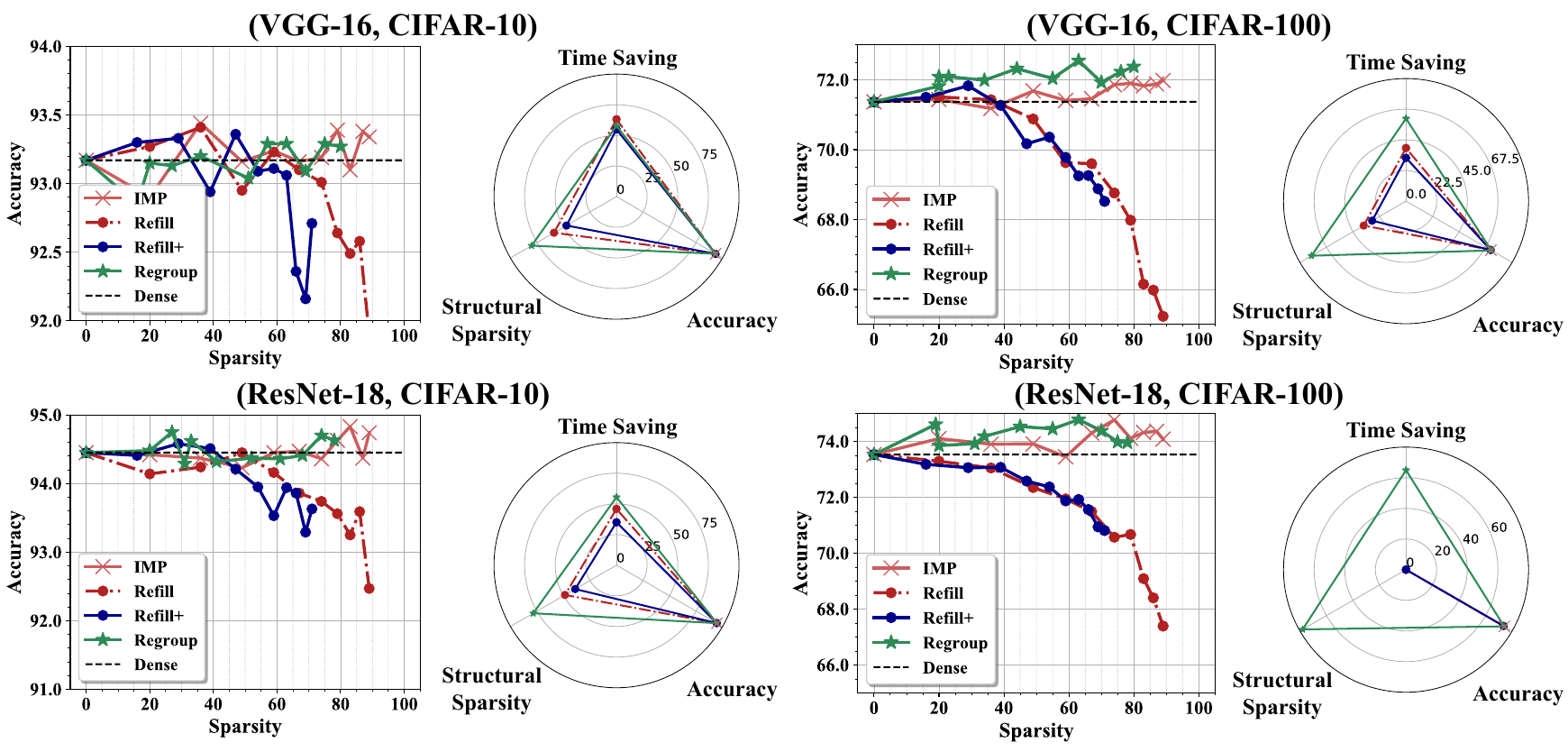}
\vspace{-8mm}
\caption{{\small (\textit{Curve plots}) Testing accuracy (\%) over sparsity (\%) on CIFAR-10/100 with large models VGG-16 ($14.72$ M) and RN-18 ($11.22$ M). (\textit{Radar plots}) The end-to-end inference time saving of extreme structural winning tickets. Note that unstructured subnetworks or dense models do not have structural sparsity, and thus they are plotted as dots in the axes of accuracy in the corresponding radar plot.}}
\vspace{-1em}
\label{fig:cifar_res_large}
\end{figure*}

\section{The Existence of Structural Winning Ticket} \label{sec:main_res}

\paragraph{Tiny-ImageNet and ImageNet.} In this section, we reveal the existence of our proposed structural winning tickets on ImageNet and Tiny-ImageNet with ResNet-50 backbone. Results of unstructured \texttt{IMP}, channel-wise structural \texttt{IMP-Refill(+)}, and group-wise structural \texttt{IMP-Regroup} are collected in the Figure~\ref{fig:imagenet_res}. The end-to-end inference time\footnote{TorchPerf \scalebox{0.85}{(\url{https://github.com/awwong1/torchprof})} is adopted as our tool to benchmark both the end-to-end and layer-wise running time on GPU devices.} of obtained structural winning tickets with extreme sparsity levels are presented, which is calculated on a single 2080 TI GPU with a batch size of $64$. Extreme sparsity is defined as maximum sparsity when the subnetwork has superior accuracy to its dense counterpart.

From Tiny-ImageNet results in Figure~\ref{fig:imagenet_res} (\textit{left}), several positive observations can be drawn: \ding{182} Structural winning tickets with $60\%$ channel-wise structural sparsity and $74\%$ group-wise structural sparsity are located by \texttt{IMP-Refill} and \texttt{IMP-Regroup} respectively, which validate the effectiveness of our proposals. \ding{183} Although at the high sparsity levels (i.e., $>50\%$), \texttt{IMP-Refill+} outperforms \texttt{IMP-Refill} if they are from the same unstructured IMP subnetworks. Considering the overall trade-off between channel-wise structural sparsity and accuracy, \texttt{IMP-Refill} appears a clear advantage. A possible explanation is that \textit{refilling+} seems to bring undesired channels which potentially result in a degraded performance trade-off. \ding{184} \texttt{IMP-Regroup} performs better at high sparsities. It is within expectation since fine-grained group-wise structural patterns tend to make the networks be more amenable to pruning. \ding{185} Extreme channel- / group-wise structural winning tickets with $45\%\sim50\%$ / $74\%$ sparsity from \texttt{IMP-Refill(+)} / \texttt{IMP-Regroup} achieve $57.53\%\sim61.79\%$ / $64.84\%$ GPU running time savings, without sacrificing accuracies. 

As for large-scale ImageNet experiments, the conclusion are slightly different: \ding{182} There is almost no difference between the performance of \texttt{IMP-Refill} and \texttt{IMP-Refill+}, and both can not find channel-wise structural winning tickets. But it seems to suggest our picking rule (i.e., channel weights' $\ell_1$ norm) provides a great estimation for channel importance, although it is too aggressive for ImageNet experiments. \ding{183} The group-wise structural winning ticket at $31\%$ sparsity still exist in (RN-50, ImageNet), while the low sparsity brings limited $1\%$ time savings. For a better efficiency and performance trade-off, \texttt{IMP-Regroup} is capable of locating structural subnetworks at $51\%$ / $58\%$ sparsity with $53.75\%$ / $60.23\%$ time savings and $0.33\%$ / $0.95\%$ accuracy drop.

\begin{figure*}[t] 
\centering
\vspace{-0.1em}
\includegraphics[width=1\linewidth]{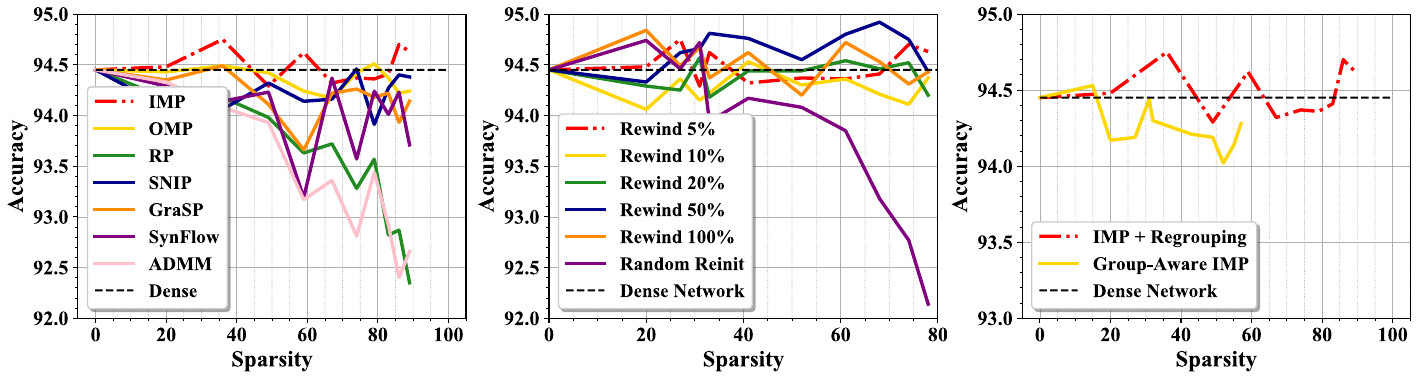}
\vspace{-8mm}
\caption{{\small (\textit{Left}) Performance of structural tickets grouped from diverse initial unstructured masks. (\textit{Middle}) Performance of group-wise structural tickets with different weight rewinding. (\textit{Right}) Performance comparisons between \texttt{IMP-Regroup} and group-aware \texttt{IMP} as described in Algorithm~\ref{alg:G_IMP}. Testing accuracies (\%) over network sparsity levels (\%) are reported on (RN-18,C10).}}
\vspace{-2mm}
\label{fig:source_res}
\end{figure*}

\vspace{-2mm}
\paragraph{CIFAR with diverse network architectures.} We then validate our approaches on CIFAR-10/100 (C10/100) with diverse network backbones including Wide-ResNet-32-2, MobileNet-v1, VGG-16, and ResNet-18. Based on the extensive results in Figure~\ref{fig:cifar_res_small} and~\ref{fig:cifar_res_large}, we find: \ding{182} On \{(WRN-32-2,C10), (WRN-32-2,C100), (MBNet-v1,C10), (MBNet-v1,C100), (VGG-16,C10), (VGG-16,C100), (RN-18,C10), (RN-18,C100)\} schemes, we consistently disclose the existence of structural winning tickets with \{$53\%$, $28\%$, $67\%$, $0\%$, $60\%$, $40\%$, $50\%$, $0\%$\} channel-wise sparsity and \{$66\%$, $36\%$, $72\%$, $56\%$, $80\%$, $80\%$, $78\%$, $78\%$\} group-wise sparsity from \texttt{IMP-Refill(+)} and \texttt{IMP-Regroup}, respectively. \ding{183} With the same network, pursuing channel-wise sparse patterns on CIFAR-100 is more challenging than it on CIFAR-10, possibly due to the larger dataset complexity. On the same dataset, larger networks tend to have larger extreme sparsities for both channel- and group-wise structural winning tickets, with the exception of \texttt{IMP-Refill(+)} on (RN-18, C100). \ding{184} At the middle sparsity levels (i.e., $<50\%$), \texttt{IMP-Regroup} behaves closely to \texttt{IMP-Refill(+)}, while \texttt{IMP-Regroup} has a superior performance at high sparsity levels. \ding{185} Up to \{$57.75\%$, $60.60\%$, $55.45\%$, $64.93\%$\} GPU running time savings are obtained by group-wise structural winning tickets with undamaged performance on \{(VGG-16,C10), (VGG-16,C100), (RN-18,C10), (RN-18,C100)\}, which surpass \texttt{IMP}, \texttt{IMP-Refill(+)}, and dense models by a significant efficiency margin. A exception is that \texttt{IMP-Refill} on (VGG-16,C10) achieves the best time savings, i.e., $63.11\%$.

\begin{figure}[!ht] 
\centering
\vspace{-0.1em}
\includegraphics[width=1\linewidth]{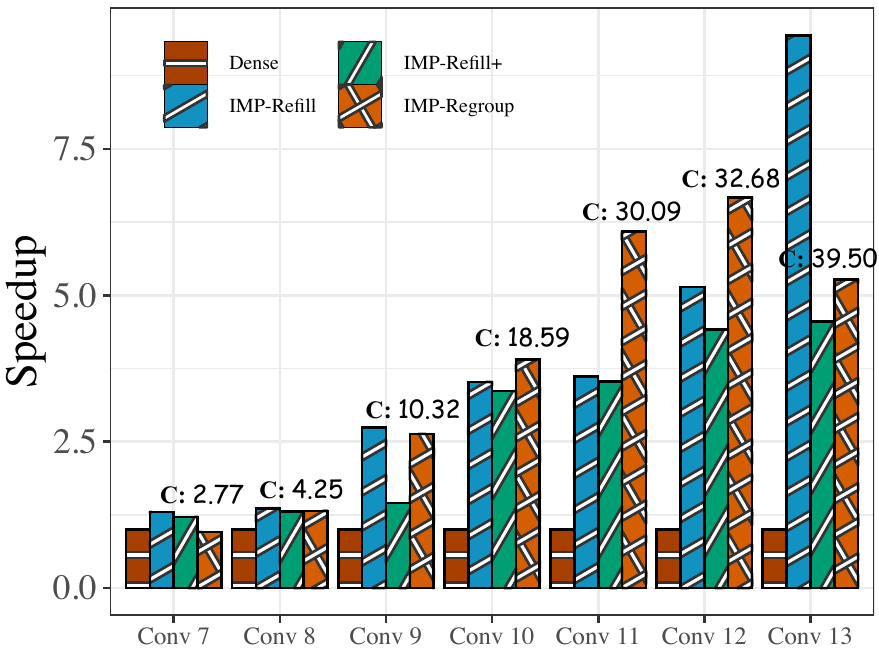}
\vspace{-8mm}
\caption{{\small The layer-wise performance of convolution operations in extreme structural winning tickets of (VGG-16, C10). The first six conv. operations are omitted since there is no meaningful speedup, coincided with~\citet{rumi2020accelerating}. Marks like ``C: 2.77" indicate the layer-wise compression ratio of \texttt{IMP-Regroup}.}}
\vspace{-2mm}
\label{fig:layerwise}
\end{figure}

\vspace{-2mm}
\paragraph{Layer-wise speedups.} Figure~\ref{fig:layerwise} and~\ref{fig:layerwise_c100} shows the layer-wise speedup performance of convolution operations in VGG-16's extreme structured winning tickets from different algorithms.\texttt{IMP-Regroup} presents impressive layer-wise speedups up to $6.67$x compared to others, especially on the last a few layers (e.g., conv. $12$). The possible reasons lie in two aspects: ($i$) the latter layers reach a larger compression ratio and have greater potentials for acceleration; ($ii$) the \textit{regrouping} algorithm prefers convolutional layers (i.e., latter layers in VGG-16) with a larger number of kernels which benefits to group appropriate dense blocks, as also suggested by~\citet{rumi2020accelerating}.

\section{Ablation Study and Visualization}

\vspace{-1mm}
\paragraph{Different sources of unstructured masks.} Intuitively, the initial unstructured sparse mask should plays an essential role in the achievable performance of our proposed ``post-processing techniques". We therefore conduct a comprehensive ablation study about the various sources of the initial sparse masks in Figure~\ref{fig:source_res}, including \texttt{IMP}, \texttt{OMP}, \texttt{RP}, \texttt{SNIP}, \texttt{GraSP}, \texttt{SynFlow}, and \texttt{ADMM}. The details of comparison methods are in Section~\ref{sec: preliminaries}. We observe that \texttt{IMP} and \texttt{OMP} provide initial unstructured masks with the top-$2$ highest quality for our \textit{regrouping} algorithm, in terms of the train-from-scratch accuracy of grouped structural subnetworks.

\vspace{-1mm}
\paragraph{Different initialization for the re-training.} Initialization~\citep{frankle2018the,Renda2020Comparing} as another key factor in LTH, also contributes significantly to the existence of winning tickets. To exhaustively investigate the effect from different initialization (e.g., rewound weights), we launch experiments started from diverse rewound weights ($\{5\%, 10\%,20\%, 50\%, 100\%\}$ of total training epochs) as well as a random re-initialization. In Figure~\ref{fig:source_res}, using $50\%$ rewound weight reaches the overall best performance; other weight rewinding setups perform similarly and clearly surpass random re-initializing at sparsity levels $>30\%$. 

\vspace{-1mm}
\paragraph{Group-aware \texttt{IMP}.} This work mainly focuses on the post-processing of unstructured sparse masks. Another possibility is integrating \textit{regrouping} into \texttt{IMP} by alternatively performing unstructured magnitude pruning and regrouping, which we term as group-aware \texttt{IMP}. From Fig.~\ref{fig:source_res}, it has a worse performance due to the stricter constraint on sparse patterns, compared to \texttt{IMP-Regroup}.  

\vspace{-1mm}
\paragraph{Extra study.} More investigations about ($1$) transfer tickets and training efficiency; ($2$) comparison with random tickets; ($3$) ablation on different training settings; ($4$) FLOPs saving; ($5$) visualization of sparse masks are in Appendix~\ref{sec:more_results}.

\vspace{-1mm}
\section{Conclusion}
\vspace{-1mm}
In this paper, we challenge the ``common sense" that an identified \texttt{IMP} winning ticket can only have unstructured sparsity, which severely limits its practical usage due to the irregular patterns. We for the first time demonstrate the existence of structural winning tickets by leveraging post-processing techniques, i.e., \textit{refilling(+)} and \textit{regrouping}. The located channel- and group-wise structural subnetworks achieve significant inference speedups up to $6.67$x on hardware platforms. In this sense, our positive results bridge the gap between the lottery ticket hypothesis and practical accelerations in real-world scenarios. We would be interested in examining LTH with more effective structural sparsity for real-time mobile computing in future work. 

\vspace{-0.2em}
\section*{Acknowledgment}
\vspace{-1mm}
Z.W. is in part supported by an NSF EPCN project (\#2053272). Y.W. is in part supported by an NSF CMMI project (\#2013067).


\bibliography{SLTH}

\begin{thebibliography}{70}
\providecommand{\natexlab}[1]{#1}
\providecommand{\url}[1]{\texttt{#1}}
\expandafter\ifx\csname urlstyle\endcsname\relax
  \providecommand{\doi}[1]{doi: #1}\else
  \providecommand{\doi}{doi: \begingroup \urlstyle{rm}\Url}\fi

\bibitem[Alabdulmohsin et~al.(2021)Alabdulmohsin, Markeeva, Keysers, and
  Tolstikhin]{alabdulmohsin2021generalized}
Alabdulmohsin, I., Markeeva, L., Keysers, D., and Tolstikhin, I.
\newblock A generalized lottery ticket hypothesis.
\newblock \emph{arXiv preprint arXiv:2107.06825}, 2021.

\bibitem[Bartlett et~al.(2021)Bartlett, Montanari, and
  Rakhlin]{bartlett2021deep}
Bartlett, P.~L., Montanari, A., and Rakhlin, A.
\newblock Deep learning: a statistical viewpoint.
\newblock \emph{arXiv preprint arXiv:2103.09177}, 2021.

\bibitem[Bartoldson et~al.(2019)Bartoldson, Morcos, Barbu, and
  Erlebacher]{bartoldson2019generalization}
Bartoldson, B.~R., Morcos, A.~S., Barbu, A., and Erlebacher, G.
\newblock The generalization-stability tradeoff in neural network pruning.
\newblock \emph{arXiv preprint arXiv:1906.03728}, 2019.

\bibitem[Blalock et~al.(2020)Blalock, Ortiz, Frankle, and
  Guttag]{blalock2020state}
Blalock, D., Ortiz, J. J.~G., Frankle, J., and Guttag, J.
\newblock What is the state of neural network pruning?
\newblock \emph{arXiv preprint arXiv:2003.03033}, 2020.

\bibitem[Brown et~al.(2020)Brown, Mann, Ryder, Subbiah, Kaplan, Dhariwal,
  Neelakantan, Shyam, Sastry, Askell, et~al.]{brown2020language}
Brown, T.~B., Mann, B., Ryder, N., Subbiah, M., Kaplan, J., Dhariwal, P.,
  Neelakantan, A., Shyam, P., Sastry, G., Askell, A., et~al.
\newblock Language models are few-shot learners.
\newblock \emph{arXiv preprint arXiv:2005.14165}, 2020.

\bibitem[Chen et~al.(2018)Chen, Oh, Fan, and Pistoia]{chen2018sc}
Chen, C.-F., Oh, J., Fan, Q., and Pistoia, M.
\newblock Sc-conv: Sparse-complementary convolution for efficient model
  utilization on cnns.
\newblock In \emph{2018 IEEE International Symposium on Multimedia (ISM)}, pp.\
   97--100. IEEE, 2018.

\bibitem[Chen et~al.(2020{\natexlab{a}})Chen, Frankle, Chang, Liu, Zhang,
  Carbin, and Wang]{chen2020lottery2}
Chen, T., Frankle, J., Chang, S., Liu, S., Zhang, Y., Carbin, M., and Wang, Z.
\newblock The lottery tickets hypothesis for supervised and self-supervised
  pre-training in computer vision models.
\newblock \emph{arXiv preprint arXiv:2012.06908}, 2020{\natexlab{a}}.

\bibitem[Chen et~al.(2020{\natexlab{b}})Chen, Frankle, Chang, Liu, Zhang, Wang,
  and Carbin]{chen2020lottery}
Chen, T., Frankle, J., Chang, S., Liu, S., Zhang, Y., Wang, Z., and Carbin, M.
\newblock The lottery ticket hypothesis for pre-trained bert networks.
\newblock \emph{arXiv preprint arXiv:2007.12223}, 2020{\natexlab{b}}.

\bibitem[Chen et~al.(2021{\natexlab{a}})Chen, Cheng, Gan, Liu, and
  Wang]{chen2021ultra}
Chen, T., Cheng, Y., Gan, Z., Liu, J., and Wang, Z.
\newblock Ultra-data-efficient gan training: Drawing a lottery ticket first,
  then training it toughly.
\newblock \emph{arXiv preprint arXiv:2103.00397}, 2021{\natexlab{a}}.

\bibitem[Chen et~al.(2021{\natexlab{b}})Chen, Sui, Chen, Zhang, and
  Wang]{chen2021unified}
Chen, T., Sui, Y., Chen, X., Zhang, A., and Wang, Z.
\newblock A unified lottery ticket hypothesis for graph neural networks,
  2021{\natexlab{b}}.

\bibitem[Chen(2018)]{chen2018escort}
Chen, X.
\newblock Escort: Efficient sparse convolutional neural networks on gpus.
\newblock \emph{arXiv preprint arXiv:1802.10280}, 2018.

\bibitem[Chen et~al.(2021{\natexlab{c}})Chen, Chen, Zhang, and
  Wang]{chen2021you}
Chen, X., Chen, T., Zhang, Z., and Wang, Z.
\newblock You are caught stealing my winning lottery ticket! making a lottery
  ticket claim its ownership.
\newblock \emph{Advances in Neural Information Processing Systems}, 34,
  2021{\natexlab{c}}.

\bibitem[Chen et~al.(2021{\natexlab{d}})Chen, Zhang, Sui, and
  Chen]{chen2021gans}
Chen, X., Zhang, Z., Sui, Y., and Chen, T.
\newblock {\{}GAN{\}}s can play lottery tickets too.
\newblock In \emph{International Conference on Learning Representations},
  2021{\natexlab{d}}.
\newblock URL \url{https://openreview.net/forum?id=1AoMhc_9jER}.

\bibitem[Chetlur et~al.(2014)Chetlur, Woolley, Vandermersch, Cohen, Tran,
  Catanzaro, and Shelhamer]{chetlur2014cudnn}
Chetlur, S., Woolley, C., Vandermersch, P., Cohen, J., Tran, J., Catanzaro, B.,
  and Shelhamer, E.
\newblock cudnn: Efficient primitives for deep learning.
\newblock \emph{arXiv preprint arXiv:1410.0759}, 2014.

\bibitem[Deng et~al.(2009)Deng, Dong, Socher, Li, Li, and
  Fei-Fei]{deng2009imagenet}
Deng, J., Dong, W., Socher, R., Li, L.-J., Li, K., and Fei-Fei, L.
\newblock Imagenet: A large-scale hierarchical image database.
\newblock In \emph{2009 IEEE conference on computer vision and pattern
  recognition}, pp.\  248--255. Ieee, 2009.

\bibitem[Dong et~al.(2019)Dong, Liu, Zhao, Li, Li, Wang, and
  Feng]{dong2019acorns}
Dong, X., Liu, L., Zhao, P., Li, G., Li, J., Wang, X., and Feng, X.
\newblock Acorns: A framework for accelerating deep neural networks with input
  sparsity.
\newblock In \emph{2019 28th International Conference on Parallel Architectures
  and Compilation Techniques (PACT)}, pp.\  178--191. IEEE, 2019.

\bibitem[Du et~al.(2018)Du, Zhai, Poczos, and Singh]{du2018gradient}
Du, S.~S., Zhai, X., Poczos, B., and Singh, A.
\newblock Gradient descent provably optimizes over-parameterized neural
  networks.
\newblock \emph{arXiv preprint arXiv:1810.02054}, 2018.

\bibitem[Frankle \& Carbin(2019)Frankle and Carbin]{frankle2018the}
Frankle, J. and Carbin, M.
\newblock The lottery ticket hypothesis: Finding sparse, trainable neural
  networks.
\newblock In \emph{International Conference on Learning Representations}, 2019.
\newblock URL \url{https://openreview.net/forum?id=rJl-b3RcF7}.

\bibitem[Frankle et~al.(2020{\natexlab{a}})Frankle, Dziugaite, Roy, and
  Carbin]{frankle2020linear}
Frankle, J., Dziugaite, G.~K., Roy, D., and Carbin, M.
\newblock Linear mode connectivity and the lottery ticket hypothesis.
\newblock In \emph{International Conference on Machine Learning}, pp.\
  3259--3269. PMLR, 2020{\natexlab{a}}.

\bibitem[Frankle et~al.(2020{\natexlab{b}})Frankle, Schwab, and
  Morcos]{Frankle2020The}
Frankle, J., Schwab, D.~J., and Morcos, A.~S.
\newblock The early phase of neural network training.
\newblock In \emph{International Conference on Learning Representations},
  2020{\natexlab{b}}.
\newblock URL \url{https://openreview.net/forum?id=Hkl1iRNFwS}.

\bibitem[Gale et~al.(2019)Gale, Elsen, and Hooker]{gale2019state}
Gale, T., Elsen, E., and Hooker, S.
\newblock The state of sparsity in deep neural networks.
\newblock \emph{arXiv preprint arXiv:1902.09574}, 2019.

\bibitem[Gan et~al.(2021)Gan, Chen, Li, Chen, Cheng, Wang, and
  Liu]{gan2021playing}
Gan, Z., Chen, Y.-C., Li, L., Chen, T., Cheng, Y., Wang, S., and Liu, J.
\newblock Playing lottery tickets with vision and language.
\newblock \emph{arXiv preprint arXiv:2104.11832}, 2021.

\bibitem[Han et~al.(2015{\natexlab{a}})Han, Mao, and Dally]{han2015deep}
Han, S., Mao, H., and Dally, W.~J.
\newblock Deep compression: Compressing deep neural networks with pruning,
  trained quantization and huffman coding.
\newblock \emph{arXiv preprint arXiv:1510.00149}, 2015{\natexlab{a}}.

\bibitem[Han et~al.(2015{\natexlab{b}})Han, Pool, Tran, and
  Dally]{han2015learning}
Han, S., Pool, J., Tran, J., and Dally, W.~J.
\newblock Learning both weights and connections for efficient neural networks.
\newblock \emph{arXiv preprint arXiv:1506.02626}, 2015{\natexlab{b}}.

\bibitem[Han et~al.(2016)Han, Liu, Mao, Pu, Pedram, Horowitz, and
  Dally]{han2016eie}
Han, S., Liu, X., Mao, H., Pu, J., Pedram, A., Horowitz, M.~A., and Dally,
  W.~J.
\newblock Eie: efficient inference engine on compressed deep neural network.
\newblock In \emph{ISCA}, 2016.

\bibitem[Hanson \& Pratt(1988)Hanson and Pratt]{hanson1988comparing}
Hanson, S. and Pratt, L.
\newblock Comparing biases for minimal network construction with
  back-propagation.
\newblock \emph{Advances in neural information processing systems}, 1:\penalty0
  177--185, 1988.

\bibitem[He et~al.(2016)He, Zhang, Ren, and Sun]{he2016deep}
He, K., Zhang, X., Ren, S., and Sun, J.
\newblock Deep residual learning for image recognition.
\newblock In \emph{Proceedings of the IEEE conference on computer vision and
  pattern recognition}, pp.\  770--778, 2016.

\bibitem[He et~al.(2017)He, Zhang, and Sun]{he2017channel}
He, Y., Zhang, X., and Sun, J.
\newblock Channel pruning for accelerating very deep neural networks.
\newblock In \emph{Proceedings of the IEEE International Conference on Computer
  Vision (ICCV)}, pp.\  1389--1397, 2017.

\bibitem[Hoefler et~al.(2021)Hoefler, Alistarh, Ben-Nun, Dryden, and
  Peste]{hoefler2021sparsity}
Hoefler, T., Alistarh, D., Ben-Nun, T., Dryden, N., and Peste, A.
\newblock Sparsity in deep learning: Pruning and growth for efficient inference
  and training in neural networks.
\newblock \emph{arXiv preprint arXiv:2102.00554}, 2021.

\bibitem[Hong et~al.(2018)Hong, Sukumaran-Rajam, Bandyopadhyay, Kim, Kurt,
  Nisa, Sabhlok, \c{C}ataly\"{u}rek, Parthasarathy, and
  Sadayappan]{hong2018efficient}
Hong, C., Sukumaran-Rajam, A., Bandyopadhyay, B., Kim, J., Kurt, S.~E., Nisa,
  I., Sabhlok, S., \c{C}ataly\"{u}rek, U.~V., Parthasarathy, S., and
  Sadayappan, P.
\newblock Efficient sparse-matrix multi-vector product on gpus.
\newblock Association for Computing Machinery, 2018.

\bibitem[Hong et~al.(2019)Hong, Sukumaran-Rajam, Nisa, Singh, and
  Sadayappan]{hong2019adaptive}
Hong, C., Sukumaran-Rajam, A., Nisa, I., Singh, K., and Sadayappan, P.
\newblock Adaptive sparse tiling for sparse matrix multiplication.
\newblock In \emph{Proceedings of the 24th Symposium on Principles and Practice
  of Parallel Programming}, pp.\  300--314, 2019.

\bibitem[Howard et~al.(2017)Howard, Zhu, Chen, Kalenichenko, Wang, Weyand,
  Andreetto, and Adam]{howard2017mobilenets}
Howard, A.~G., Zhu, M., Chen, B., Kalenichenko, D., Wang, W., Weyand, T.,
  Andreetto, M., and Adam, H.
\newblock Mobilenets: Efficient convolutional neural networks for mobile vision
  applications.
\newblock \emph{arXiv preprint arXiv:1704.04861}, 2017.

\bibitem[Hu et~al.(2016)Hu, Peng, Tai, and Tang]{hu2016network}
Hu, H., Peng, R., Tai, Y.-W., and Tang, C.-K.
\newblock Network trimming: A data-driven neuron pruning approach towards
  efficient deep architectures.
\newblock \emph{arXiv preprint arXiv:1607.03250}, 2016.

\bibitem[Jiang et~al.(2020)Jiang, Hong, and Agrawal]{10.1145/3332466.3374546}
Jiang, P., Hong, C., and Agrawal, G.
\newblock A novel data transformation and execution strategy for accelerating
  sparse matrix multiplication on gpus.
\newblock PPoPP '20. Association for Computing Machinery, 2020.

\bibitem[Kaplan et~al.(2020)Kaplan, McCandlish, Henighan, Brown, Chess, Child,
  Gray, Radford, Wu, and Amodei]{kaplan2020scaling}
Kaplan, J., McCandlish, S., Henighan, T., Brown, T.~B., Chess, B., Child, R.,
  Gray, S., Radford, A., Wu, J., and Amodei, D.
\newblock Scaling laws for neural language models.
\newblock \emph{arXiv preprint arXiv:2001.08361}, 2020.

\bibitem[Krizhevsky et~al.(2009)Krizhevsky, Hinton,
  et~al.]{krizhevsky2009learning}
Krizhevsky, A., Hinton, G., et~al.
\newblock Learning multiple layers of features from tiny images.
\newblock 2009.

\bibitem[Le \& Yang(2015)Le and Yang]{le2015tiny}
Le, Y. and Yang, X.
\newblock Tiny imagenet visual recognition challenge.
\newblock \emph{CS 231N}, 7:\penalty0 7, 2015.

\bibitem[LeCun et~al.(1990)LeCun, Denker, and Solla]{lecun1990optimal}
LeCun, Y., Denker, J.~S., and Solla, S.~A.
\newblock Optimal brain damage.
\newblock In \emph{Advances in neural information processing systems}, pp.\
  598--605, 1990.

\bibitem[Lee et~al.(2019{\natexlab{a}})Lee, Ajanthan, and Torr]{lee2018snip}
Lee, N., Ajanthan, T., and Torr, P.
\newblock Snip: Single-shot network pruning based on connection sensitivity.
\newblock In \emph{International Conference on Learning Representations
  (ICLR)}, 2019{\natexlab{a}}.

\bibitem[Lee et~al.(2019{\natexlab{b}})Lee, Ajanthan, and Torr]{snip}
Lee, N., Ajanthan, T., and Torr, P.
\newblock Snip: Single-shot network pruning based on connection sensitivity.
\newblock In \emph{International Conference on Learning Representations},
  2019{\natexlab{b}}.
\newblock URL \url{https://openreview.net/forum?id=B1VZqjAcYX}.

\bibitem[Li et~al.(2016)Li, Kadav, Durdanovic, Samet, and Graf]{li2016pruning}
Li, H., Kadav, A., Durdanovic, I., Samet, H., and Graf, H.~P.
\newblock Pruning filters for efficient convnets.
\newblock \emph{arXiv preprint arXiv:1608.08710}, 2016.

\bibitem[Lin et~al.(2020)Lin, Ji, Zhang, Zhang, Wu, and Tian]{lin2020channel}
Lin, M., Ji, R., Zhang, Y., Zhang, B., Wu, Y., and Tian, Y.
\newblock Channel pruning via automatic structure search.
\newblock \emph{arXiv preprint arXiv:2001.08565}, 2020.

\bibitem[Liu et~al.(2017)Liu, Li, Shen, Huang, Yan, and Zhang]{liu2017learning}
Liu, Z., Li, J., Shen, Z., Huang, G., Yan, S., and Zhang, C.
\newblock Learning efficient convolutional networks through network slimming.
\newblock In \emph{Proceedings of the IEEE International Conference on Computer
  Vision}, pp.\  2736--2744, 2017.

\bibitem[Louizos et~al.(2017)Louizos, Welling, and Kingma]{louizos2017learning}
Louizos, C., Welling, M., and Kingma, D.~P.
\newblock Learning sparse neural networks through $ l\_0 $ regularization.
\newblock \emph{arXiv preprint arXiv:1712.01312}, 2017.

\bibitem[Ma et~al.(2021{\natexlab{a}})Ma, Chen, Hu, You, Xie, and
  Wang]{ma2021good}
Ma, H., Chen, T., Hu, T.-K., You, C., Xie, X., and Wang, Z.
\newblock Good students play big lottery better.
\newblock \emph{arXiv preprint arXiv:2101.03255}, 2021{\natexlab{a}}.

\bibitem[Ma et~al.(2020)Ma, Guo, Niu, Lin, Tang, Ma, Ren, and
  Wang]{ma2020pconv}
Ma, X., Guo, F.-M., Niu, W., Lin, X., Tang, J., Ma, K., Ren, B., and Wang, Y.
\newblock Pconv: The missing but desirable sparsity in dnn weight pruning for
  real-time execution on mobile devices.
\newblock In \emph{Proceedings of the AAAI Conference on Artificial
  Intelligence}, volume~34, pp.\  5117--5124, 2020.

\bibitem[Ma et~al.(2021{\natexlab{b}})Ma, Yuan, Shen, Chen, Chen, Chen, Liu,
  Qin, Liu, Wang, et~al.]{ma2021sanity}
Ma, X., Yuan, G., Shen, X., Chen, T., Chen, X., Chen, X., Liu, N., Qin, M.,
  Liu, S., Wang, Z., et~al.
\newblock Sanity checks for lottery tickets: Does your winning ticket really
  win the jackpot?
\newblock \emph{arXiv preprint arXiv:2107.00166}, 2021{\natexlab{b}}.

\bibitem[Mao et~al.(2017)Mao, Han, Pool, Li, Liu, Wang, and
  Dally]{mao2017exploring}
Mao, H., Han, S., Pool, J., Li, W., Liu, X., Wang, Y., and Dally, W.~J.
\newblock Exploring the regularity of sparse structure in convolutional neural
  networks.
\newblock \emph{arXiv preprint arXiv:1705.08922}, 2017.

\bibitem[Molchanov et~al.(2019)Molchanov, Mallya, Tyree, Frosio, and
  Kautz]{molchanov2019importance}
Molchanov, P., Mallya, A., Tyree, S., Frosio, I., and Kautz, J.
\newblock Importance estimation for neural network pruning.
\newblock In \emph{Proceedings of the IEEE Conference on Computer Vision and
  Pattern Recognition}, pp.\  11264--11272, 2019.

\bibitem[Niu et~al.(2020)Niu, Ma, Lin, Wang, Qian, Lin, Wang, and
  Ren]{niu2020patdnn}
Niu, W., Ma, X., Lin, S., Wang, S., Qian, X., Lin, X., Wang, Y., and Ren, B.
\newblock Patdnn: Achieving real-time dnn execution on mobile devices with
  pattern-based weight pruning.
\newblock \emph{arXiv preprint arXiv:2001.00138}, 2020.

\bibitem[Park et~al.(2016)Park, Li, Wen, Tang, Li, Chen, and
  Dubey]{park2016faster}
Park, J., Li, S., Wen, W., Tang, P. T.~P., Li, H., Chen, Y., and Dubey, P.
\newblock Faster cnns with direct sparse convolutions and guided pruning.
\newblock In \emph{International Conference on Learning Representations}, 2016.

\bibitem[Peng et~al.(2017)Peng, Fu, Liu, and Hsu]{peng2017adaptive}
Peng, K.-Y., Fu, S.-Y., Liu, Y.-P., and Hsu, W.-C.
\newblock Adaptive runtime exploiting sparsity in tensor of deep learning
  neural network on heterogeneous systems.
\newblock In \emph{2017 International Conference on Embedded Computer Systems:
  Architectures, Modeling, and Simulation (SAMOS)}, pp.\  105--112. IEEE, 2017.

\bibitem[Radu et~al.(2019)Radu, Kaszyk, Wen, Turner, Cano, Crowley, Franke,
  Storkey, and O'Boyle]{radu2019performance}
Radu, V., Kaszyk, K., Wen, Y., Turner, J., Cano, J., Crowley, E.~J., Franke,
  B., Storkey, A., and O'Boyle, M.
\newblock Performance aware convolutional neural network channel pruning for
  embedded gpus.
\newblock In \emph{2019 IEEE International Symposium on Workload
  Characterization (IISWC)}, pp.\  24--34. IEEE, 2019.

\bibitem[Ren et~al.(2018)Ren, Zhang, Ye, Li, Xu, Qian, Lin, and
  Wang]{ren2018admmnn}
Ren, A., Zhang, T., Ye, S., Li, J., Xu, W., Qian, X., Lin, X., and Wang, Y.
\newblock Admm-nn: An algorithm-hardware co-design framework of dnns using
  alternating direction method of multipliers, 2018.

\bibitem[Renda et~al.(2020)Renda, Frankle, and Carbin]{Renda2020Comparing}
Renda, A., Frankle, J., and Carbin, M.
\newblock Comparing rewinding and fine-tuning in neural network pruning.
\newblock In \emph{8th International Conference on Learning Representations},
  2020.

\bibitem[Ruder(2016)]{ruder2016overview}
Ruder, S.
\newblock An overview of gradient descent optimization algorithms.
\newblock \emph{arXiv preprint arXiv:1609.04747}, 2016.

\bibitem[Rumi et~al.(2020)Rumi, Ma, Wang, and Jiang]{rumi2020accelerating}
Rumi, M.~A., Ma, X., Wang, Y., and Jiang, P.
\newblock Accelerating sparse cnn inference on gpus with performance-aware
  weight pruning.
\newblock In \emph{Proceedings of the ACM International Conference on Parallel
  Architectures and Compilation Techniques}, pp.\  267--278, 2020.

\bibitem[Shangguan et~al.(2019)Shangguan, Li, Liang, Alvarez, and
  McGraw]{shangguan2019optimizing}
Shangguan, Y., Li, J., Liang, Q., Alvarez, R., and McGraw, I.
\newblock Optimizing speech recognition for the edge.
\newblock \emph{arXiv preprint arXiv:1909.12408}, 2019.

\bibitem[Simonyan \& Zisserman(2014)Simonyan and Zisserman]{simonyan2014very}
Simonyan, K. and Zisserman, A.
\newblock Very deep convolutional networks for large-scale image recognition.
\newblock \emph{arXiv preprint arXiv:1409.1556}, 2014.

\bibitem[Tanaka et~al.(2020)Tanaka, Kunin, Yamins, and
  Ganguli]{tanaka2020pruning}
Tanaka, H., Kunin, D., Yamins, D.~L., and Ganguli, S.
\newblock Pruning neural networks without any data by iteratively conserving
  synaptic flow.
\newblock In \emph{Advances in Neural Information Processing Systems 33
  pre-proceedings}, 2020.

\bibitem[Tang et~al.(2020)Tang, Wang, Xu, Tao, Xu, Xu, and Xu]{tang2020scop}
Tang, Y., Wang, Y., Xu, Y., Tao, D., Xu, C., Xu, C., and Xu, C.
\newblock Scop: Scientific control for reliable neural network pruning.
\newblock \emph{Advances in Neural Information Processing Systems},
  33:\penalty0 10936--10947, 2020.

\bibitem[Wang et~al.(2020)Wang, Zhang, and Grosse]{Wang2020Picking}
Wang, C., Zhang, G., and Grosse, R.
\newblock Picking winning tickets before training by preserving gradient flow.
\newblock In \emph{International Conference on Learning Representations}, 2020.
\newblock URL \url{https://openreview.net/forum?id=SkgsACVKPH}.

\bibitem[Wen et~al.(2016)Wen, Wu, Wang, Chen, and Li]{wen2016learning}
Wen, W., Wu, C., Wang, Y., Chen, Y., and Li, H.
\newblock Learning structured sparsity in deep neural networks.
\newblock In \emph{Advances in neural information processing systems
  (NeurIPS)}, pp.\  2074--2082, 2016.

\bibitem[You et~al.(2020)You, Li, Xu, Fu, Wang, Chen, Baraniuk, Wang, and
  Lin]{You2020Drawing}
You, H., Li, C., Xu, P., Fu, Y., Wang, Y., Chen, X., Baraniuk, R.~G., Wang, Z.,
  and Lin, Y.
\newblock Drawing early-bird tickets: Toward more efficient training of deep
  networks.
\newblock In \emph{International Conference on Learning Representations}, 2020.
\newblock URL \url{https://openreview.net/forum?id=BJxsrgStvr}.

\bibitem[Yu et~al.(2020)Yu, Edunov, Tian, and Morcos]{yu2019playing}
Yu, H., Edunov, S., Tian, Y., and Morcos, A.~S.
\newblock Playing the lottery with rewards and multiple languages: lottery
  tickets in rl and nlp.
\newblock In \emph{8th International Conference on Learning Representations},
  2020.

\bibitem[Zagoruyko \& Komodakis(2016)Zagoruyko and
  Komodakis]{zagoruyko2016wide}
Zagoruyko, S. and Komodakis, N.
\newblock Wide residual networks.
\newblock \emph{arXiv preprint arXiv:1605.07146}, 2016.

\bibitem[Zhang et~al.(2018)Zhang, Ye, Zhang, Tang, Wen, Fardad, and
  Wang]{Zhang_2018}
Zhang, T., Ye, S., Zhang, K., Tang, J., Wen, W., Fardad, M., and Wang, Y.
\newblock A systematic {DNN} weight pruning framework using alternating
  direction method of multipliers.
\newblock In \emph{ECCV}, 2018.

\bibitem[Zhang et~al.(2021)Zhang, Chen, Chen, and Wang]{pmlr-v139-zhang21c}
Zhang, Z., Chen, X., Chen, T., and Wang, Z.
\newblock Efficient lottery ticket finding: Less data is more.
\newblock In Meila, M. and Zhang, T. (eds.), \emph{Proceedings of the 38th
  International Conference on Machine Learning}, volume 139 of
  \emph{Proceedings of Machine Learning Research}, pp.\  12380--12390. PMLR,
  18--24 Jul 2021.
\newblock URL \url{https://proceedings.mlr.press/v139/zhang21c.html}.

\bibitem[Zhou et~al.(2021)Zhou, Ma, Zhu, Liu, Zhang, Yuan, Sun, and
  Li]{zhou2021learning}
Zhou, A., Ma, Y., Zhu, J., Liu, J., Zhang, Z., Yuan, K., Sun, W., and Li, H.
\newblock Learning n: M fine-grained structured sparse neural networks from
  scratch.
\newblock \emph{arXiv preprint arXiv:2102.04010}, 2021.

\bibitem[Zhu et~al.(2019)Zhu, Zhang, Gu, and Xie]{Zhu2019STC}
Zhu, M., Zhang, T., Gu, Z., and Xie, Y.
\newblock Sparse tensor core: Algorithm and hardware co-design for vector-wise
  sparse neural networks on modern gpus.
\newblock In \emph{Proceedings of the 52nd Annual IEEE/ACM International
  Symposium on Microarchitecture}, pp.\  359--371, 2019.

\end{thebibliography}
\bibliographystyle{icml2022}

\newpage
\appendix
\renewcommand{\thepage}{A\arabic{page}}  
\renewcommand{\thesection}{A\arabic{section}}   
\renewcommand{\thetable}{A\arabic{table}}   
\renewcommand{\thefigure}{A\arabic{figure}}

\clearpage

\section{More Implementation Details} \label{sec:more_implementation}

\paragraph{Group-aware \texttt{IMP}.} Here we provides the detailed procedures of group-aware \texttt{IMP} in Algorithm~\ref{alg:G_IMP}. Intuitively, it embeds \textit{regrouping} (Algorithm~\ref{alg:regroup}) into \texttt{IMP} (Algorithm~\ref{alg:IMP}) by performing \textit{regrouping} on the unstructured mask $m$ from each \texttt{IMP} round. 

\begin{minipage}{0.48\textwidth}
\begin{algorithm}[H]
\caption{Group-aware \texttt{IMP}} \label{alg:G_IMP}
\begin{algorithmic}[1]
\REQUIRE {$f(x;\theta_0)$, group-wise structural sparsity $s$}
\ENSURE {$f(x; m\odot\theta_i)$ with group-wise structural sparse mask $s$}
\STATE Set the pruning mask $m=\boldsymbol{1}\in\mathbb R^{|\theta|}$\\
Train $f(x;\theta_0)$ to rewinding step $i$: $f(x;\theta_i)=\mathcal{A}_i^{\mathcal{T}}(f(x;\theta_0))$  \\
\WHILE {not reach sparsity $s$}{
\STATE Train $f(x;m\odot\theta_i)$ to step $t$: $f(x;m\odot\theta_t)=\mathcal{A}_{t-i}^{\mathcal{T}}(f(x;m\odot\theta_i))$ \\
\STATE Pruning $20\%$ of remaining weight of $m\odot\theta_t$, and update $m$ \\
\STATE Refining the unstructured mask $m$ by performing $\textit{regrouping}$, as shown in Algorithm~\ref{alg:regroup}}
\ENDWHILE 
\end{algorithmic}
\end{algorithm}
\end{minipage}

\paragraph{Profiling.} To compute the GPU running time of regrouped convolution layers, we adopt their CUDA C/C++ implementation. Our results do not include the running time of normalization and activation layers, following the standard in \citet{rumi2020accelerating}. For a fair calculation, we feed the \textit{same} input features to convolution layers that belong to the same model. For ResNet-18 and VGG-16, the size of the input features is $(64, 64, 127, 127)$. For ResNet-50, the size of input features is $(64, 64, 64, 64)$. The GPU we use for profiling is NVIDIA RTX 2080 TI, with a CUDA version of $10.2$ and a cuDNN~\citep{chetlur2014cudnn} version of $7.6.5$. 

\section{More Experiment Results} \label{sec:more_results}

\begin{figure}[t] 
\centering
\includegraphics[width=1\linewidth]{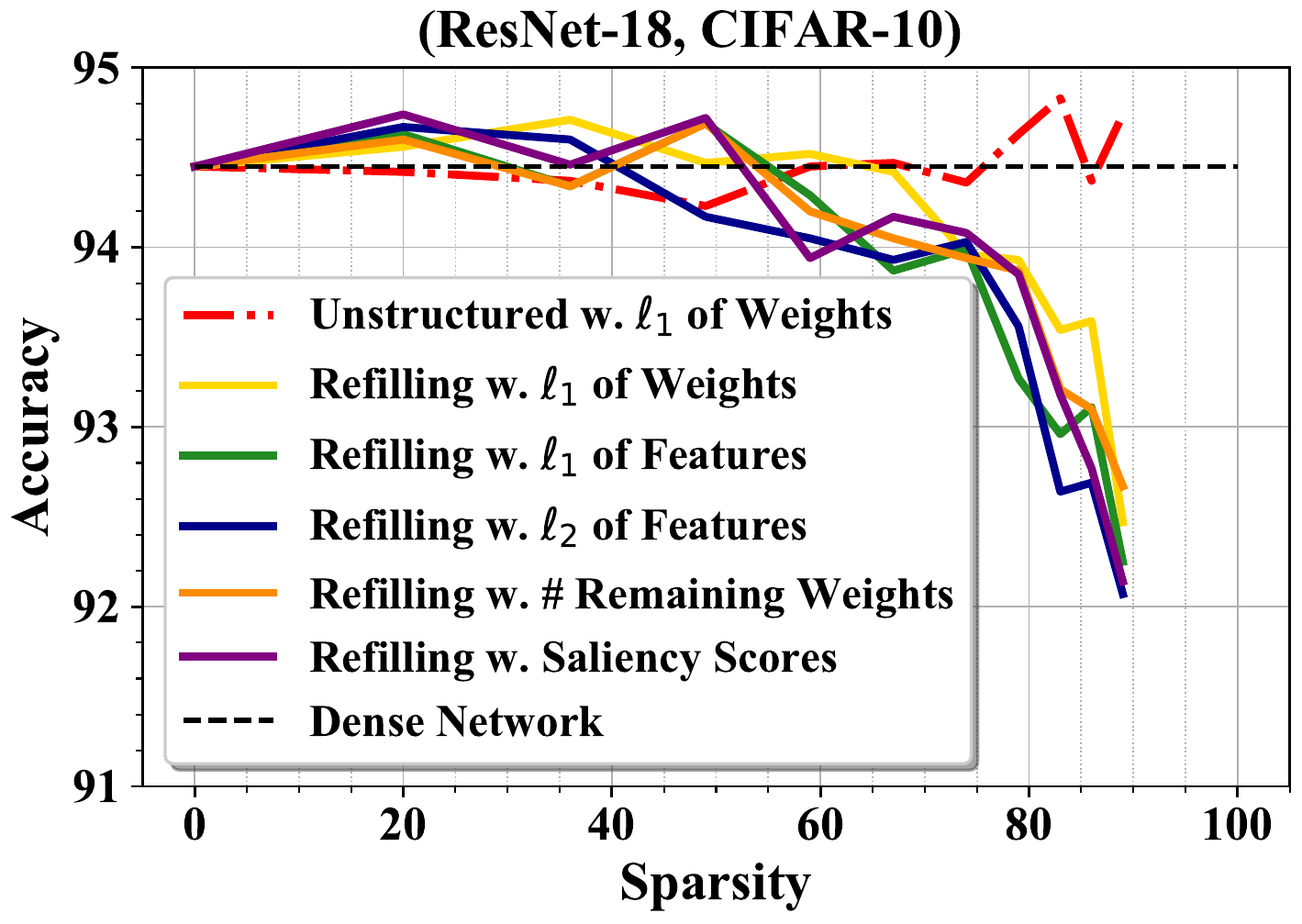}
\vspace{-6mm}
\caption{{\small Performance of structural tickets refilled by diverse channel picking criterion. Testing accuracies (\%) over network sparsity levels (\%) are reported on (RN-18,C10).}}
\vspace{-4mm}
\label{fig:pick_rules}
\end{figure}

\paragraph{Different channel picking criterion for refilling.} We ablation the channel picking criterion for \texttt{IMP-Refill(+)}, including \ding{182} the $\ell_1$ norm of channel's remaining weight, \ding{183} the $\ell_1$ or $\ell_2$ norms of channel's feature map, \ding{184} the number of remaining weights in the channel, \ding{185} the channel's saliency score~\citep{molchanov2019importance}. Experiment results are collected in Figure~\ref{fig:pick_rules}, which demonstrate the superior performance of \texttt{IMP-Refill} w. $\ell_1$ of channel weights (\textcolor{yellow}{yellow} curve in Figure~\ref{fig:pick_rules}).

\paragraph{Transfer tickets and training efficiency.} We investigate the transferability of our found (fine-grained) structural winning tickets, which grants the extra bonus of training efficiency to our proposals. Specifically, following the setups in~\citet{chen2020lottery2}, we first identify refilled and regrouped structural winning tickets in ResNet-18 on ImageNet, and then transfer them to the downstream CIFAR-10 task. Transfer results are presented in Table~\ref{tab:new_pretrain}. Compared to the dense network baseline ($95.37\%$), \texttt{IMP-Refill} locates channel-wise structural (transfer) winning tickets at the sparsity around $36\%$, and \texttt{IMP-Regroup} locates group-wise structural (transfer) winning tickets at a higher sparsity (more than $56.00\%$). Such an encouraging transfer study means that we can even replace the full model with a much smaller subnetwork while maintaining an undamaged downstream performance. And this is also why our \texttt{IMP-Regroup} and \texttt{IMP-Refill} obtain $7.14\%\sim 34.67\%$ and $34.53\%$ \textbf{training time savings} during downstream training with matched or even improved generalization. Note that this efficient training is an extra benefits of our proposal, in addition to impressive inference efficiency. 

\begin{table}[!htbp]
\centering
\caption{Transfer accuracy ($\%$), time saving ($\%$) and remaining weights ($\%$) on ResNet-50 with CIFAR-10. \texttt{IMP-Refill} and \texttt{IMP-Regroup} are evaluated. The baseline accuracy of dense network is $95.37\%$.}
\resizebox{1\linewidth}{!}{
\begin{tabular}{c|c|c}
\toprule
\multicolumn{3}{c}{\texttt{IMP-Refill}}  \\ \midrule
Remaining Weight & Transfer Accuracy & Time Savings \\ \midrule
64.14\% & 95.81\% & 34.53\% \\ 
51.37\% & 95.14\% & 48.10\% \\ 
41.01\% & 94.51\% & 60.67\% \\ 
32.76\% & 94.38\% & 65.98\% \\
26.17\% & 94.19\% & 69.04\% \\
20.97\% & 94.11\% & 71.08\% \\
\toprule
\multicolumn{3}{c}{\texttt{IMP-Regroup}} \\ \midrule
Remaining Weight & Transfer Accuracy & Time Savings \\ \midrule
59.43\% & 95.65\% & 7.14\% \\
51.84\% & 95.39\% & 21.85\% \\
43.99\% & 95.51\% & 34.67\% \\
\bottomrule
\end{tabular}}
\label{tab:new_pretrain}
\end{table}

\paragraph{Comparison with random tickets.} As a sanity check, we conduct a comparison with random tickets~\citep{frankle2018the} which are trained from random re-initialization. Experiments results on (RN-18, C10) are collected in Table~\ref{tab:random_1} and~\ref{tab:random_2}. We show that random tickets have obviously inferior performance, which suggests that our identified refilled and regrouped subnetworks are highly non-trivial (fine-grained) structural winning tickets.

\begin{table}[!htbp]
\vspace{-4mm}
\caption{Testing accuracy ($\%$) of $\texttt{IMP-Refill}$ with rewinding weights (Ours) and random re-initialization (Random Tickets).}
\label{tab:random_1}
\centering
\resizebox{1\linewidth}{!}{
\begin{tabular}{c|c|c|c}
\toprule
IMP Round & Remaining Weight & Accuracy (Ours) & Accuracy (Random Tickets) \\
\midrule
1 & 80.29$\%$ & 94.14\% & 94.04\% \\
2 & 64.49$\%$ & 94.24\% & 93.96\% \\
3 & 51.43$\%$ & 94.45\% & 94.20\% \\
4 & 41.24$\%$ & 94.16\% & 93.98\% \\
5 & 32.97$\%$ & 93.86\% & 93.53\% \\
\bottomrule
\end{tabular}}
\vspace{-4mm}
\end{table}

\begin{table}[!htbp]
\vspace{-4mm}
\caption{Testing accuracy ($\%$) of $\texttt{IMP-Regroup}$ with rewinding weights (Ours) and random re-initialization (Random Tickets).}
\label{tab:random_2}
\centering
\resizebox{1\linewidth}{!}{
\begin{tabular}{c|c|c|c}
\toprule
IMP Round & Remaining Weight & Accuracy (Ours) & Accuracy (Random Tickets)  \\
\midrule
1 & 80.00$\%$ & 94.48\% & 94.19\% \\
2 & 72.95$\%$ & 94.75\% & 93.42\% \\
3 & 69.37$\%$ & 94.29\% & 93.58\% \\
4 & 67.12$\%$ & 94.62\% & 93.84\% \\
5 & 58.54$\%$ & 94.32\% & 93.76\% \\
\bottomrule
\end{tabular}}
\vspace{-4mm}
\end{table}

\paragraph{Different training settings.} To validate our algorithm's effectiveness under different training configurations, we perform extra experiments with VGG-16(+), WRN-32-2(+), and RN-50(+). The changes of training settings are summarized as below:
\begin{enumerate}[leftmargin=*]
\item [\ding{172}] For VGG-16(+), we increase the number of training epochs to $240$, and decay the learning rate at $150$-th, $180$-th, and $210$-th epoch.
\item [\ding{173}] For WRN-32-2(+), we do not split the official training set into the a training and a validation set as our other experiments did. We also report the best validation accuracy instead of the best test accuracy. The number of training epochs is increased to $240$ and the learning rate is decayed at $150$-th, $180$-th, and $210$-th epoch.
\item [\ding{174}] For RN-50(+), we replace the first convolution layer to be of kernel size $3$, padding size $1$, and strides $1$.
\end{enumerate}

$\rhd$ \textit{VGG-16(+) on C100}. As shown in Table~\ref{tab:new_vgg16_cf100}, we demonstrated that our conclusions are still hold: \texttt{IMP-Regroup} can locate structural winning tickets at very high sparsity levels (e.g., $>75\%$).  

\begin{table}[t]
\centering
\vspace{-2mm}
\caption{Testing accuracy ($\%$) and remaining weights ($\%$) on CIFAR-100 with VGG-16(+). \texttt{IMP}, \texttt{IMP-Refill}, and \texttt{IMP-Regroup} are evaluated. The baseline accuracy of dense network is $73.43\%$.}
\resizebox{1\linewidth}{!}{
\begin{tabular}{c|c|c|c|c|c|c}
\toprule
\multirow{2}{*}{Round} & \multicolumn{2}{c|}{\texttt{IMP}} & \multicolumn{2}{c|}{\texttt{IMP-Refill}} & \multicolumn{2}{c}{\texttt{IMP-Regroup}} \\
\cmidrule{2-7}
& Remaining Weight & Accuracy & Remaining Weight & Accuracy & Remaining Weight & Accuracy \\
\midrule
1 & 80.00\% & 73.64 & 80.17\% & 73.43 & 82.36\% & 73.63 \\
2 & 64.00\% & 73.80 & 64.06\% & 72.87 & 80.00\% & 73.81 \\
3 & 51.20\% & 73.67 & 51.31\% & 72.67 & 69.46\% & 74.31 \\
4 & 40.96\% & 74.01 & 41.08\% & 71.37 & 62.61\% & 73.94 \\
5 & 32.77\% & 74.27 & 32.85\% & 70.79 & 56.09\% & 75.05 \\
6 & 26.21\% & 74.56 & 26.33\% & 71.07 & 46.53\% & 74.98 \\
7 & 20.97\% & 74.58 & 21.03\% & 69.42 & 38.18\% & 75.24 \\
8 & 16.78\% & 74.52 & 16.94\% & 68.75 & 30.98\% & 74.68 \\
9 & 13.42\% & 74.42 & 13.42\% & 67.25 & 25.27\% & 75.25 \\
\bottomrule
\end{tabular}}
\vspace{-4mm}
\label{tab:new_vgg16_cf100}
\end{table}

$\rhd$ \textit{WRN-32-2(+) on C100.} As shown in Table~\ref{tab:new_wrn32_cf100}, we find consistent observations: \texttt{IMP-Regroup} locates structural winning tickets at about $75\%$ sparsity, and \texttt{IMP-Refill} identifies structural winning tickets at $20\%$ sparsity. 

\begin{table}[t]
\centering
\caption{Testing accuracy ($\%$) and remaining weights ($\%$) on CIFAR-100 with WideResNet-32-2(+). \texttt{IMP}, \texttt{IMP-Refill}, and \texttt{IMP-Regroup} are evaluated. The baseline accuracy of dense network is $75.53\%$.}
\resizebox{1\linewidth}{!}{
\begin{tabular}{c|c|c|c|c|c|c}
\toprule
\multirow{2}{*}{Round} & \multicolumn{2}{c|}{\texttt{IMP}} & \multicolumn{2}{c|}{\texttt{IMP-Refill}} & \multicolumn{2}{c}{\texttt{IMP-Regroup}} \\
\cmidrule{2-7}
& Remaining Weight & Accuracy & Remaining Weight & Accuracy & Remaining Weight & Accuracy \\
\midrule
1 & 80.00\% & 76.21 & 80.00\% & 75.46 & 80.00\% & 75.98 \\
2 & 64.00\% & 75.78 & 64.06\% & 74.59 & 64.00\% & 76.19 \\
3 & 51.20\% & 76.02 & 51.51\% & 73.53 & 51.20\% & 76.13 \\
4 & 40.96\% & 75.92 & 41.51\% & 72.95 & 40.96\% & 75.88 \\
6 & 26.21\% & 75.74 & 26.53\% & 70.91 & 26.27\% & 75.78 \\
7 & 20.97\% & 75.92 & 21.11\% & 69.55 & 21.76\% & 74.74 \\
8 & 16.78\% & 75.87 & 17.11\% & 67.74 & 18.14\% & 73.85 \\
9 & 13.42\% & 75.41 & 13.67\% & 65.73 & 14.85\% & 72.99 \\
\bottomrule
\end{tabular}}
\vspace{-3mm}
\label{tab:new_wrn32_cf100}
\end{table}

\begin{table}[t]
\centering
\vspace{-2mm}
\caption{Testing accuracy ($\%$) and remaining weights ($\%$) on Tiny-ImageNet with ResNet-50(+). \texttt{IMP}, \texttt{IMP-Refill}, and \texttt{IMP-Regroup} are evaluated. The baseline accuracy of dense network is $65.33\%$.}
\resizebox{1\linewidth}{!}{
\begin{tabular}{c|c|c|c|c|c|c}
\toprule
\multirow{2}{*}{Round} & \multicolumn{2}{c|}{\texttt{IMP}} & \multicolumn{2}{c|}{\texttt{IMP-Refill}} & \multicolumn{2}{c}{\texttt{IMP-Regroup}} \\
\cmidrule{2-7}
& Remaining Weight & Accuracy & Remaining Weight & Accuracy & Remaining Weight & Accuracy \\
\midrule
1 & 80.00\% & 65.44 & 80.30\% & 65.27 & 80.15\% & 65.51 \\
2 & 64.00\% & 65.69 & 64.16\% & 63.40 & 68.25\% & 65.16 \\
3 & 51.20\% & 65.50 & 51.42\% & 61.89 & 58.19\% & 65.21 \\
4 & 40.96\% & 65.73 & 41.08\% & 60.43 & 54.19\% & 64.42 \\
5 & 32.77\% & 65.23 & 32.85\% & 59.64 & 51.75\% & 64.52 \\
\bottomrule
\end{tabular}}
\vspace{-4mm}
\label{tab:new_res50_ti}
\end{table}

\begin{figure}[!htb]
\centering
\includegraphics[width=0.95\linewidth]{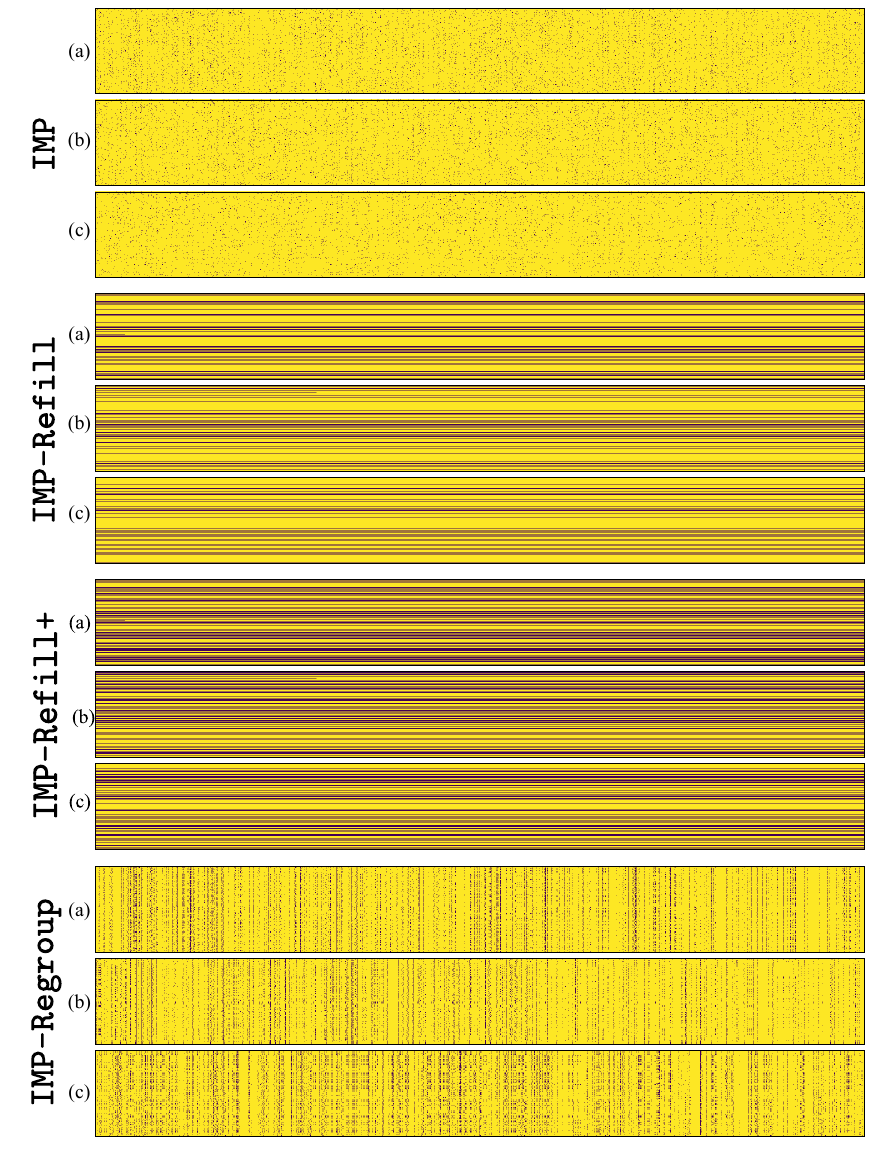}
\vspace{-1.1em}
\caption{{\small Sparse mask visualizations of the extreme winning tickets from \texttt{IMP} (unstructured), \texttt{IMP-Refill(+)} (channel-wise structural), and \texttt{IMP-Regroup} (group-wise structural) on (VGG-16,C10). The darker color indicates the remaining unpruned elements. (a,b,c) are the last three conv. layers.}}
\vspace{-4mm}
\label{fig:mask_vis}
\end{figure}

$\rhd$ \textit{RN-50(+) on Tiny-ImageNet.} Experimental results in Table~\ref{tab:new_res50_ti} suggest that: \texttt{IMP-Regroup} locates structural winning tickets at about $42\%$ sparsity, and \texttt{IMP-Refill} discovers structural winning tickets at $20\%$ sparsity, which echo our findings in the main text. 

\vspace{-2mm}
\paragraph{FLOPs saving.} For a sufficient evaluation, we calculate the FLOPs of diverse subnetworks from VGG-16 on CIFAR-10 dataset. The FLOPs of a dense VGG-16 is about $0.314$G. We select sparsity levels across different methods as similar as possible for a better comparison. Subnetworks from \texttt{IMP-Refill}, \texttt{IMP-Refill+}, and \texttt{IMP-Regroup} at sparsity levels of \{$32.84\%$, $46.41\%$, $20.12\%$\} have \{$0.089$G, $0.122$G, $0.093$G\} FLOPs, respectively. It is noteworthy that \texttt{Refill} and \texttt{Refill+} trim down the input and output channels of a convolution layer while Regroup cannot. Thus, \texttt{Refill} and \texttt{Refill+} can save more FLOPs under a similar sparsity level. 

\vspace{-2mm}
\paragraph{Visualization of sparse masks.} Figure~\ref{fig:mask_vis} visualizes different types of obtained sparse masks from (VGG-16,C10). Sub-figures (a,b,c) plot the mask matrices of size $c_{\mathrm{out}}\times n$ for certain layers. Similar to the illustration in Figure~\ref{fig:methods}, \texttt{IMP-Refill(+)} masks show clear kernel-wise sparse patterns across the rows, and \texttt{IMP-Regroup} masks present fine-grained structural sparse patterns capable of forming neat dense blocks after regrouping. 

\begin{figure}[!ht] 
\centering
\vspace{-0.5em}
\includegraphics[width=0.9\linewidth]{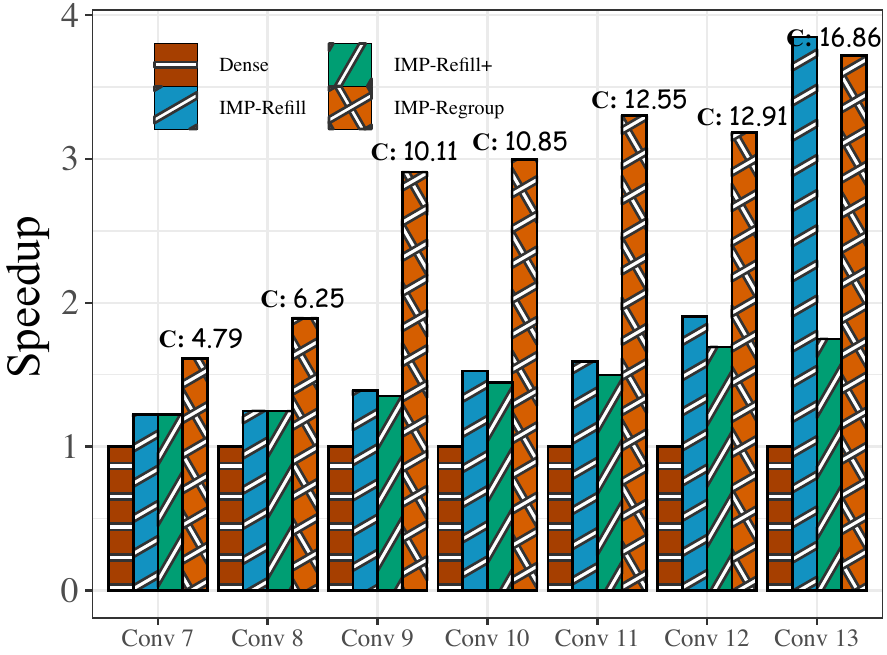}
\vspace{-4mm}
\caption{{\small The layer-wise performance of convolution operations in extreme structural winning tickets of (VGG-16, C10). The first six conv. operations are omitted since there is no meaningful speedup, coincided with~\citet{rumi2020accelerating}. Marks like ``C: 2.77" indicate the layer-wise compression ratio of \texttt{IMP-Regroup}.}}
\vspace{-3mm}
\label{fig:layerwise_c100}
\end{figure}

\vspace{-2mm}
\paragraph{More results of layer-wise speedups.} Figure~\ref{fig:layerwise_c100} presents extra layer-wise speedup results of VGG-16 on CIFAR-100. Similar observations to Figure~\ref{fig:layerwise} can be obtain.

\begin{figure}[!htbp]
\centering
\includegraphics[width=0.98\linewidth]{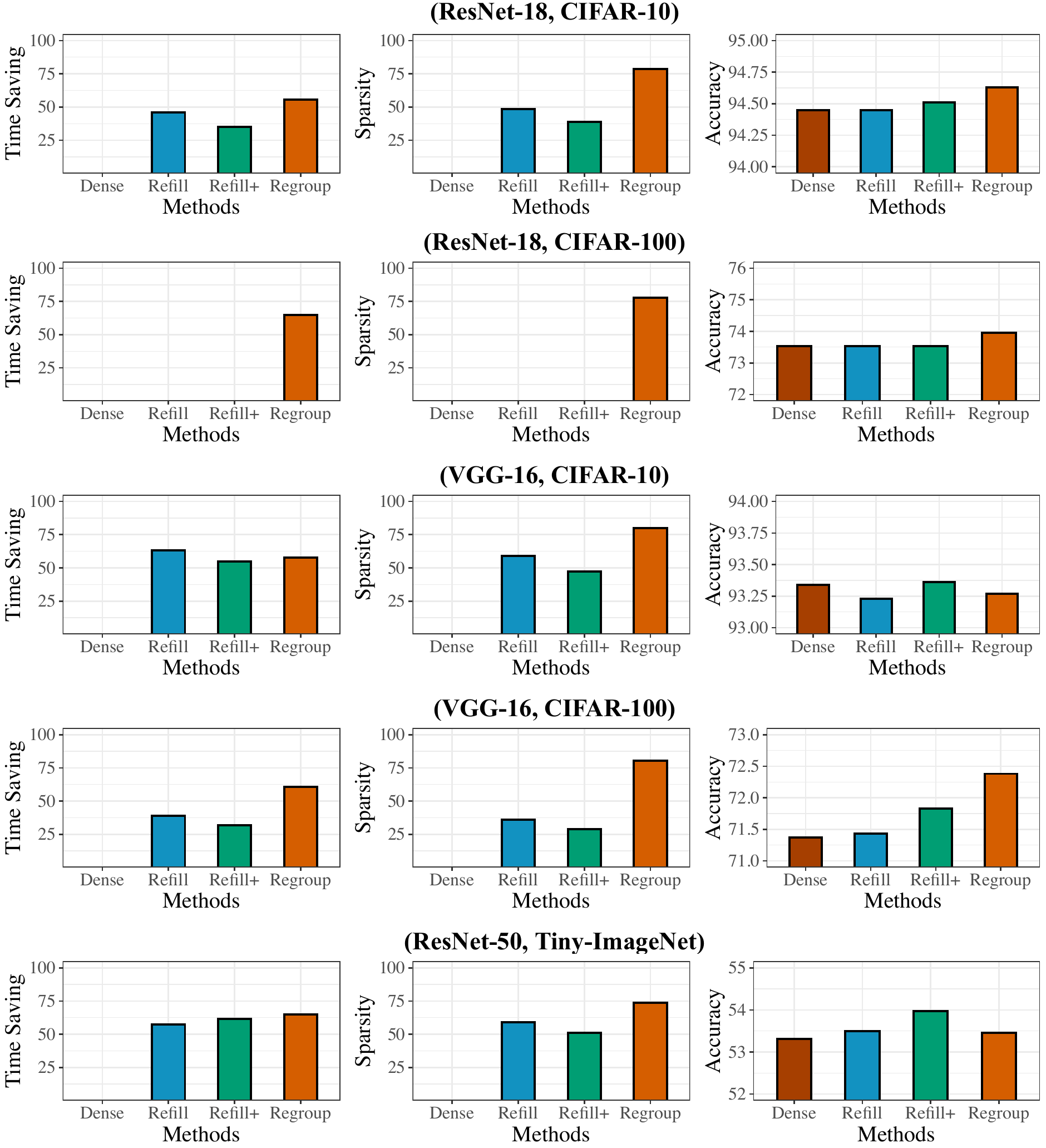}
\vspace{-4mm}
\caption{Time saving ($\%$), sparsity level ($\%$), and test accuracy ($\%$) of various models \{ResNet-18, VGG-19, ResNet-50\} on different datasets \{CIFAR-10/100, Tiny ImageNet\}.}
\vspace{-2mm}
\label{fig:histo}
\end{figure}

\vspace{-2mm}
\paragraph{Different visualizations of the radar plots.} We offer an alternative histogram visualization (Figure~\ref{fig:histo}) for radar plots in the main text. In each histogram, four approaches are reported: \texttt{Dense}, \texttt{IMP-Refill}, \texttt{IMP-Refill+}, and \texttt{IMP-Regroup}. \texttt{Dense} as the compared baseline with zero time saving, so the corresponding bars are always unseen from the charts. 

\begin{figure}[!ht]
    \centering
    \vspace{-3mm}
    \includegraphics[width=1\linewidth]{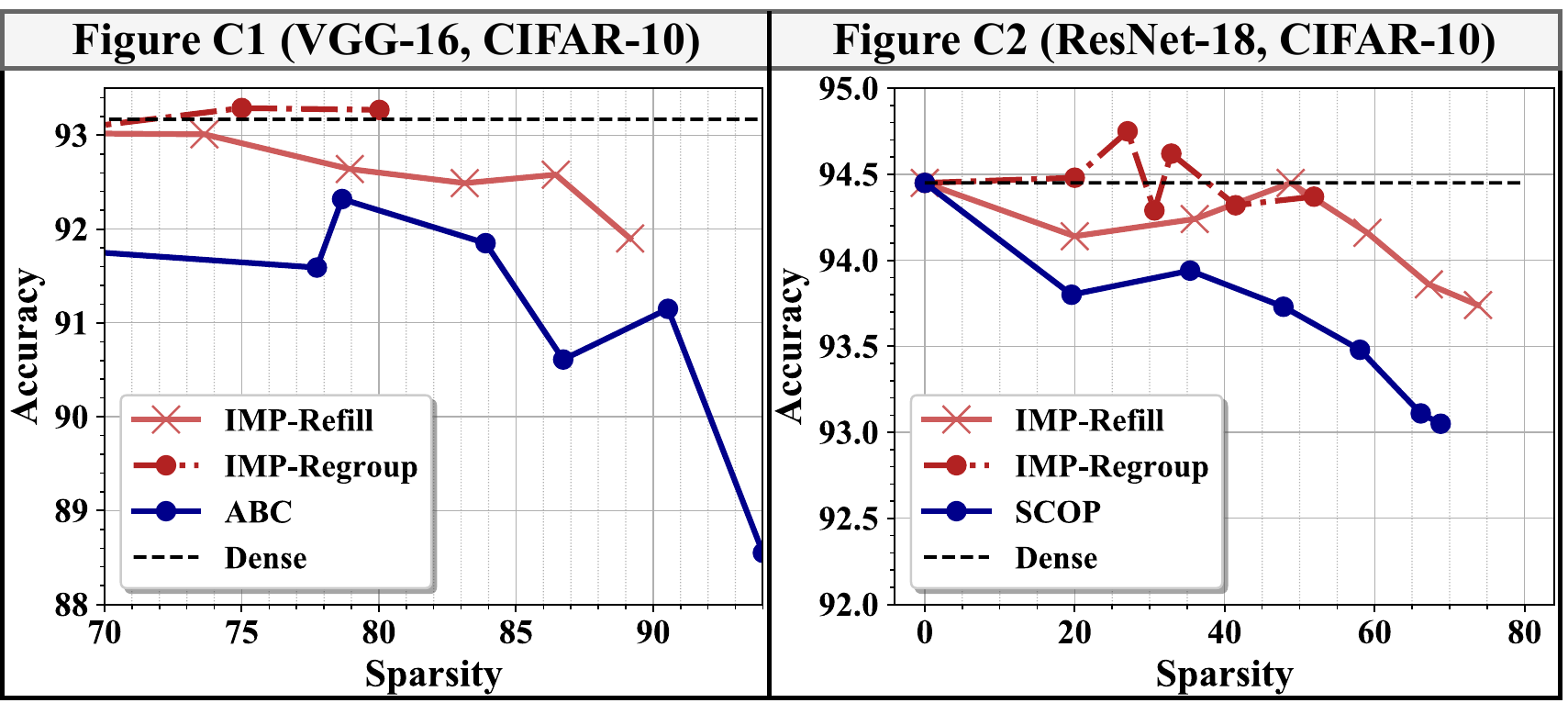}
    \vspace{-8mm}
    \caption{{\small Comparisons with NAS and SOTA structural pruning.}}
    \vspace{-9mm}
    \label{fig:nas}
\end{figure}

\vspace{-2mm}
\paragraph{Comparison to NAS and SOTA structure pruning methods.} We conduct extra experiments with a neural architecture search (NAS) based approach, i.e., ABC pruner~\citep{lin2020channel}. Specifically, it searches the channel numbers per layer, and then the derived structure will be trained from the same initialization. At the same sparsity level $77\%$, our \texttt{IMP-Refill} surpasses the ABC pruner by $1.5\%$ accuracy, which demonstrates the superiority of located refill tickets. More sparse levels are presented in Fig.~\ref{fig:nas} (C1). 

Under the same setup of training from scratch, we further compare our proposal with SCOP~\citep{tang2020scop} and observe that ours ($94.16\sim94.62\%$) outperform SCOP ($93.48\%$) by up to $1.14\%$ accuracy at $\sim41\%$ sparsity. More sparse levels are collected in Fig.~\ref{fig:nas} (C2).

\end{document}